\documentclass[10pt,twocolumn,letterpaper]{article}

\usepackage[pagenumbers]{cvpr}              %

\usepackage{graphicx}
\usepackage{amsmath}
\usepackage{amssymb}
\usepackage{booktabs}
\usepackage{microtype}  %
\usepackage[dvipsnames]{xcolor}
\usepackage[makeroom]{cancel}

\usepackage{currfile}
\usepackage[normalem]{ulem}  %
\usepackage{enumitem}  %
\usepackage{colortbl}
\usepackage{afterpage}
\usepackage{nicefrac}
\usepackage{siunitx}
\usepackage[hang,flushmargin,symbol]{footmisc}
\usepackage{blindtext}
\usepackage{multirow}
\usepackage{siunitx}
\usepackage{lipsum}
\usepackage[figuresleft]{rotating}

\usepackage[accsupp]{axessibility}  %

\usepackage[numbers,sort,compress]{natbib}
\usepackage[pagebackref,breaklinks,colorlinks]{hyperref}

\usepackage{pifont}
\renewcommand{\checkmark}{\ding{51}}
\newcommand{\crossmark}{\ding{55}}

\usepackage[capitalize]{cleveref}
\crefname{section}{Sec.}{Secs.}
\Crefname{section}{Section}{Sections}
\Crefname{table}{Table}{Tables}
\crefname{table}{Tab.}{Tabs.}

\definecolor{mydarkblue}{rgb}{0,0.08,0.55}
\hypersetup{  %
    colorlinks=true,
    linkcolor=mydarkblue,
    citecolor=mydarkblue,
    filecolor=mydarkblue,
    urlcolor=mydarkblue
}

\def\cvprPaperID{1933}
\def\confName{CVPR}
\def\confYear{2023}

\def \rust {RUST}
\def \slsr {S}
\def \oursrt {SRT$^\dagger$}
\def \ourupsrt {UpSRT$^\dagger$}
\def \latentpose {$\tilde{p}$}

\newcommand{\ipgood}{{\color{OliveGreen}$p_x$}}
\newcommand{\tpgood}{{\color{OliveGreen}$p_y$}}
\newcommand{\ipbad}{{\color{BrickRed}$\hat p_x$}}
\newcommand{\tpbad}{{\color{BrickRed}$\hat p_y$}}
\newcommand{\ipnone}{{\color{BrickRed}\cancel{$p_x$}}}
\newcommand{\tpnone}{{\color{BrickRed}\cancel{$p_y$}}}

\let\oldparagraph\paragraph
\renewcommand{\paragraph}[1]{\oldparagraph{#1.}}
\RequirePackage{xspace}

\begin{document}

\title{RUST: Latent Neural Scene Representations from Unposed Imagery}

\author{
Mehdi S. M. Sajjadi\footnotemark[1] \quad Aravindh Mahendran \quad Thomas Kipf \\
Etienne Pot \quad Daniel Duckworth \quad Mario Lučić \quad Klaus Greff \\[0.2cm]
Google Research, Brain Team\vspace{-0.1cm}}

\maketitle

\footnotetext[1]{
Correspondence: \href{mailto:rust@msajjadi.com}{rust@msajjadi.com}.
Project page: \href{https://rust-paper.github.io/}{rust-paper.github.io}.
Contributions: MS: Conception, model design, implementation lead, analysis, infrastructure, experiments, writing, project lead.
AM:  Conception, model design, implementation, code reviews, experiments-MSN, analysis-MSN, writing.
TK: RUST explicit pose estimation, model figures, scoping, advising, writing.
EP: GNeRF experiments, Street View dataset, interactive visualization for embedding inspection.
DD: COLMAP experiments, evaluation.
ML: Team buy-in, scoping, writing.
KG: Conception, early MSN analysis, analysis-SV, data-collection and analysis, dataset generation, visualizations, scoping, advising, writing.
}

\begin{abstract}
Inferring the structure of 3D scenes from 2D observations is a fundamental challenge in computer vision.
Recently popularized approaches based on neural scene representations have achieved tremendous impact and have been applied across a variety of applications.
One of the major remaining challenges in this space is training a single model which can provide latent representations which effectively generalize beyond a single scene.
Scene Representation Transformer (SRT) has shown promise in this direction, but scaling it to a larger set of diverse scenes is challenging and necessitates accurately posed ground truth data.
To address this problem, we propose \rust\ (Really Unposed Scene representation Transformer), a pose-free approach to novel view synthesis trained on RGB images alone.
Our main insight is that one can train a Pose Encoder that peeks at the target image and learns a latent pose embedding which is used by the decoder for view synthesis.
We perform an empirical investigation into the learned latent pose structure and show that it allows meaningful test-time camera transformations and accurate explicit pose readouts.
Perhaps surprisingly, \rust\ achieves similar quality as methods which have access to perfect camera pose, thereby unlocking the potential for large-scale training of amortized neural scene representations.
\end{abstract}

\vspace{-2mm}
\section{Introduction}
\label{sec:intro}
\vspace{-1mm}

Implicit neural representations have shown remarkable ability in capturing the 3D structure of complex real-world scenes while circumventing many of the downsides of mesh based, point cloud based, and voxel grid based representations~\cite{tewari2021advances}.
Apart from visually pleasing novel view synthesis~\cite{Mildenhall20eccv_nerf, MartinBrualla21cvpr_nerfw}, such representations have shown potential for semantics~\cite{nesf}, object decomposition~\cite{osrt}, and physics simulation~\cite{driess2022learning} which makes them promising candidates for applications in augmented reality and robotics.
However, to be useful in such applications, they need to (1) provide meaningful representations when conditioned on a very limited number of views, (2) have low latency for real-time rendering, and (3) produce scene representations that facilitate generalization of knowledge to novel views and scenes.
\begin{figure}[t]
\centering
\includegraphics[width=1.0\columnwidth]{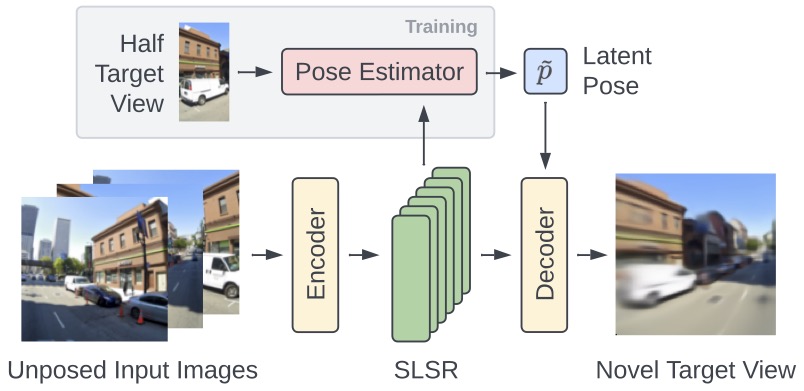}
\caption{
\textbf{Model overview} --
\rust\ produces 3D-centric scene representations through novel view synthesis purely from RGB images without requiring any camera poses.
For training, a novel \emph{Pose Estimator} module glimpses at the target view and passes a low-dimensional latent pose feature to the decoder.
\vspace{-4mm}
}
\label{fig:teaser}
\end{figure}

The recently proposed Scene Representation Transformer (SRT)~\cite{srt} exhibits most of these properties.
It achieves state-of-the-art novel view synthesis in the regime when only a handful of posed input views are available, and produces representations that are well-suited for later segmentation at both semantic~\cite{srt} and instance level~\cite{osrt}.
A major challenge in scaling methods such as SRT, however, is the difficulty in obtaining accurately posed real world data which precludes the training of the models.
We posit that it should be possible to train \emph{truly} pose-free models from RGB images alone without requiring any ground truth pose information.

We propose \rust\ (\emph{Really Unposed Scene representation Transformer}), a novel method for neural scene representation learning through novel view synthesis that does not require pose information: neither for training, nor for inference; neither for input views, nor for target views.
While at first glance it may seem unlikely that such a model could be trained, our key insight is that a sneak-peek at the target view at training can be used to infer an \emph{implicit latent pose}, thereby allowing the rendering of the correct view.
We find that our model not only learns meaningful, controllable latent pose spaces, but the quality of its novel views are even comparable to the quality of posed methods.

\vspace{2mm}
\noindent Our key contributions are as follows:
\vspace{-2mm}
\begin{itemize}[leftmargin=*]
    \setlength\itemsep{-1mm}
    
    \item We propose \rust, a novel method that learns latent 3D scene representations through novel view synthesis on very complex synthetic and real datasets without any pose information.
    \item Our model strongly outperforms prior methods in settings with noisy camera poses while matching their performance when accurate pose information is available to the baselines.
    \item We provide an investigation into the structure of the learned latent pose spaces and demonstrate that meaningful camera transformations naturally emerge in a fully unsupervised fashion.
    \item Finally, we demonstrate that the representations learned by the model allow explicit pose readout and dense semantic segmentation.
\end{itemize}

\section{Method}
\label{sec:method}

The model pipeline is shown in \cref{fig:teaser}.
A data point consists of an unordered set of $N$ \emph{input} views $x=\{x_i\in\mathbb{R}^{H \times W \times 3}\}$ of a scene.
Unlike SRT~\cite{srt}, they only consist of RGB images
since \rust\ does not use explicit poses.
Given these input views, the training objective is to predict a novel \emph{target} view $y\in\mathbb{R}^{H\times W\times 3}$ of the same scene.

To this end, the input views $x$ are first encoded using a combination of a CNN and a transformer, resulting in the Set-Latent Scene Representation (SLSR) $\slsr$ which captures the contents of the scene.
The target view $y$ is then rendered by a transformer-based decoder that attends into the SLSR $\slsr$ to retrieve relevant information about the scene for novel view synthesis.
In addition, the decoder must be conditioned on some form of a query that identifies the desired view.
Existing methods, including SRT~\cite{srt}, often use the explicit relative camera pose $p$ between one of the input views and the target view for this purpose.
This imbues an explicit notion of 3D space into the model, which introduces a burdensome requirement for accurate camera poses, especially for training such models.
\rust\ resolves this fundamental limitation by learning its own notion of implicit poses.

\begin{figure}[t]
\centering
\includegraphics[width=0.75\columnwidth]{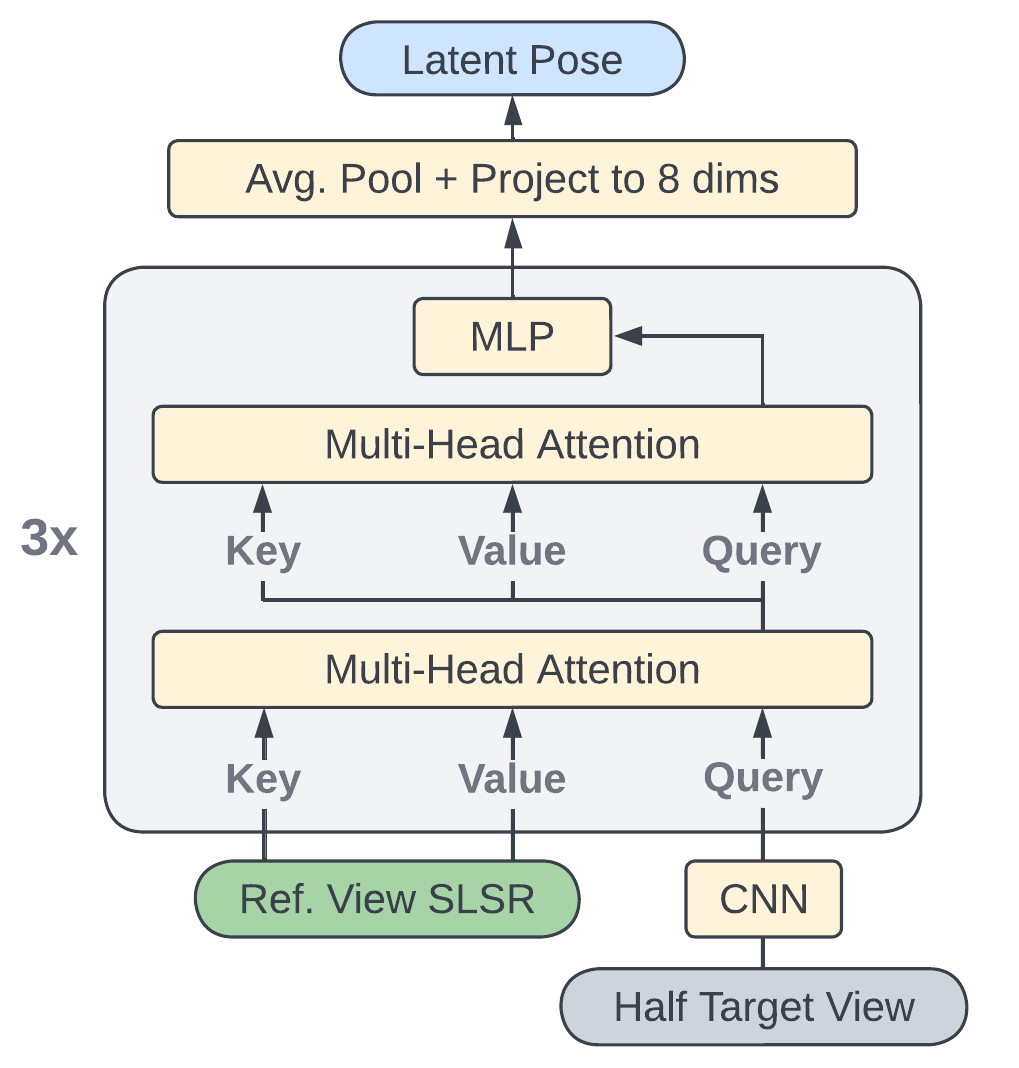}
\caption{
\textbf{Pose estimator model} --
A randomly chosen half of the target image is encoded into a latent pose.
The model mainly consists of a CNN encoder followed by alternating cross- and self-attention layers.
The final output is projected to 8 dimensions to encourage an easily controllable latent representation.
}
\vspace{-3mm}
\label{fig:architecture}
\end{figure}

\vspace{-1mm}
\paragraph{Implicit poses}
Instead of querying the decoder with explicit poses, we allow the model to learn its own implicit space of camera poses through a learned \emph{Pose Estimator} module.
For training, the Pose Estimator sees parts of the target view $y$ and the SLSR $\slsr$ and extracts a low-dimensional \emph{latent pose} feature $\tilde{p}$.
The decoder transformer then uses $\tilde{p}$ as a query to cross-attend into the SLSR to ultimately render the full novel view $\tilde{y}$.
This form of self supervision allows the model to be trained with standard reconstruction losses without requiring any pose information.
At test time, latent poses can be computed on the input views and subsequently modified for novel view synthesis, see \cref{sec:exp:lpi}.

\subsection{Model components}
\label{sec:method:components}

\paragraph{Input view encoder}
The encoder consists of a convolutional neural network (CNN) followed by a transformer.
Each input image $x_i$ is encoded independently by the shared CNN which consists of 3 downsampling blocks, each of which halves the image height and width.
As a result, each spatial output feature corresponds to an 8$\times$8 patch in the original input image.

We add the same learned position embeddings to all spatial feature maps to mark their spatial 2D position in the images.
We further add another learned embedding only to the features of the first input image $x_1$ to allow the model to distinguish them from the others.
This is relevant for the Pose Estimator module, as explained below.
Finally, we flatten all spatial feature maps and combine them across input views into a single set of tokens.
The encoder transformer then performs self-attention on this set of tokens, thereby exchanging information between the patch features.
This results in the SLSR $\slsr$ which captures the scene content as a bag of tokens.

\paragraph{Pose estimator}
\label{sec:method:poseestimator}
The model architecture of the \emph{Pose Estimator} is shown in \cref{fig:architecture}.
A randomly chosen half of the target view $y$ (\ie, either the left or the right half of the image) is first embedded into a set of tokens using a CNN similar to the input view encoder.
A transformer then alternates between cross-attending from the target view tokens into a specific subset $\tilde{\slsr}\subset\slsr$ of the SLSR and self-attending between the target view tokens.
The intuition behind cross-attending into $\tilde{\slsr}$ is that the latent pose $\tilde{p}$ should be \emph{relative} to the scene.
We allow the Pose Estimator to only attend into SLSR tokens belonging to the (arbitrarily chosen) first input view after empirical findings that this leads to better-structured latent pose spaces.
It is important to note that $\tilde{\slsr}$ contains information about \emph{all} input views due to the self-attention in the preceding encoder transformer.

Finally, we apply global mean pooling on the transformer's output and linearly project it down to an 8-dimensional latent pose $\tilde{p}$.
It is important to note that we call $\tilde{p}$ the estimated \emph{``pose''} since it primarily serves the purpose of informing the decoder of the target camera pose.
However, we do not enforce any explicit constraints on the latent, instead allowing the model to freely choose the structure of the latent poses.
Nevertheless, we find that the model learns to model meaningful camera poses, see \cref{sec:exp:lpi}.

\vspace{-1mm}
\paragraph{Decoder transformer}
\label{sec:method:decoder}

Similar to SRT, each pixel is decoded independently by a decoder transformer that cross-attends from a query into the SLSR $\slsr$, thereby aggregating relevant information from the scene representation to establish the appearance of the novel view point.
We initialize the query by concatenating the latent pose $\tilde{p}$ with the spatial 2D position of the pixel in the target image and passing it through a small query MLP.
The output of the decoder is the single RGB value of the target pixel.

\vspace{1mm}
\subsection{Training procedure}
\label{sec:method:training}

For each data point during training, we use 5 input views and 3 novel target views.
The training objective is to render the target views by minimizing the mean-squared error between the predicted output and the target view:
$||\tilde{y} - y||_2^2$.
The entire model is trained end-to-end using the Adam optimizer~\cite{adam}.
In practice, we found the model to perform better when gradients flowing to and through the Pose Estimator module are scaled down by $0.2$ which is inspired by spatial transformer networks~\cite{stn}.

\section{Related works}
\label{sec:related}

\paragraph{Neural rendering}
The field of neural rendering is vast and the introduction of Neural Radiance Fields (NeRF)~\cite{Mildenhall20eccv_nerf} has led to a surge of recent follow-up works.
NeRF optimizes an MLP to map 3D positions of the scene to radiance and density values, which are used through a differentiable volumetric rendering equation to reconstruct the provided posed images of the scene.
We refer the reader to surveys on neural rendering~\cite{tewari2021advances} and NeRF~\cite{gao2022nerf} for recent overviews.

\paragraph{NeRF without pose}
Most methods based on NeRF assume the availability of perfect camera pose.
There exist a number of works extending NeRF for pose estimation or to no-pose settings.
INeRF~\cite{YenChen20iros_iNeRF} uses pre-trained NeRFs to estimate the camera pose of novel views of the same scene.
However, posed imagery is required for each scene in order to obtain the original NeRF models.
NeRF$\text{-}\text{-}$~\cite{wang2021nerfmm} jointly optimizes the NeRF MLP along with the camera poses for the images.
BARF~\cite{Lin21iccv_BARF} adds a progressive position encoding scheme for improved gradients.
Both methods only succeed for forward-facing datasets while failing for more complex camera distributions unless noisy initial poses are given.
GNeRF~\cite{Meng21iccv_GNeRF} works for scenes with more complex camera distributions, however, the prior distribution must be known for sampling.
VMRF~\cite{vmrf} extends this to settings where the prior distribution is unknown.

All methods above require a comparably large number of images of the same scene, since NeRF tends to fail with few observations even with perfect pose~\cite{Yu21cvpr_pixelNeRF}.
While methods exist that optimize NeRFs from fewer observations~\cite{Jain21iccv_DietNeRF, niemeyer2022regnerf}, they have limitations in terms of scene complexity~\cite{srt} and to the best of our knowledge, no NeRF-based method has been demonstrated to work with few unposed images.

\vspace{-2mm}
\paragraph{Latent 3D scene representations}
In the right setup, NeRF produces high-quality novel views, though it does so without providing tangible scene representations that could be readily used for downstream tasks.
The line of research focusing on latent 3D scene representations includes NeRF-VAE~\cite{Kosiorek21icml_NeRF_VAE} which uses a variational autoencoder~\cite{vae} to learn a generative model of NeRF's for synthetic scenes, and GQN~\cite{eslami2018neural} which adds per-image latent representations to compute a global scene representation that is used by a recurrent latent variable model for novel view synthesis.
A recent extension~\cite{rosenbaum2018learning} of GQN performs pose estimation for novel target views by optimizing, at inference time, the posterior probability over poses as estimated by the generative model.
All the methods above require ground truth poses for training and inference.
GIRAFFE~\cite{Niemeyer21cvpr_GIRAFFE} requires no poses for training, but a prior distribution of camera views for the specific dataset.
It uses a purely generative mechanism, without the ability to render novel views for given scenes.

\vspace{-2mm}
\paragraph{Posed and ``unposed'' SRT}
The current most scalable latent method for neural scene representations is the Scene Representation Transformer (SRT)~\cite{srt}.
It uses a transformer-based encoder-decoder architecture for novel view synthesis, thereby scaling to much more complex scenes than prior work.
Its scene representation, the SLSR, has been shown to be useful for 3D-centric supervised and unsupervised semantic downstream tasks~\cite{srt, osrt}.

While SRT requires posed imagery, the ``unposed'' variant UpSRT has been proposed as well in the original work~\cite{srt}.
Similar to \rust, it does not require posed \emph{input} views.
However, training an UpSRT model requires exact pose information for the \emph{target} views to specify to the decoder which exact view of the scene to render.
This restricts the application of UpSRT to settings where accurate pose information is available for the entire training dataset.
We analyze further implications of this in \cref{sec:exp:nvs}.

\paragraph{Using targets as inputs for training}
UViM~\cite{kolesnikov2022uvim} encodes targets such as panoptic segmentation maps into short latent codes during training to allow a single model architecture to perform well across several tasks.
A bottleneck is enforced on the so-called \emph{guiding codes} through discretization.
Recent work on inferring protein conformations and poses~\cite{rosenbaum2021inferring} trains a VAE model in order to encode Cryo-EM images into latent distributions over poses and protein conformations.
Unlike \rust, their method requires a base scene representation that is fixed a priori and uses an explicit differentiable rendered and explicit geometry in the architecture.

\def \y {\textcolor{green}\checkmark}
\def \n {\textcolor{red}\crossmark}

\begin{table}[t]
\centering
\begin{tabular}{lcl}
\toprule
Method & Pose & PSNR %
\\
\midrule
SRT~\cite{srt} & \ipgood,\,\tpgood & 23.31 \\
\oursrt & \ipgood,\,\tpgood & 24.40  \\
\oursrt & \ipbad,\,\tpgood & 23.81  \\
\ourupsrt & \ipnone,\,\tpgood & 23.03  \\
\oursrt & \ipbad,\,\tpbad & 18.65  \\
\ourupsrt & \ipnone,\,\tpbad & 18.64  \\
\rust\ & \ipnone,\,\tpnone & 23.49  \\
\bottomrule
\end{tabular}
\begin{tabular}{ll}
\toprule
Ablation & PSNR
\\
\midrule
Right-half PE & 23.88  \\
Stop grad. & 23.16  \\
No SLSR & 20.83  \\
No self-attn & 22.97  \\
3-dim. \latentpose\ &  20.40 \\
64-dim. \latentpose\ & 23.40  \\
768-dim. \latentpose\ & 23.11  \\
\bottomrule
\end{tabular}
\caption{
    \textbf{Quantitative results on MSN} --
    \textbf{Left:}
    Comparison with prior work in various settings: perfect (\ipgood, \tpgood), noisy (\ipbad, \tpbad) and lack of (\ipnone, \tpnone) input and target poses.
    We report SRT both as proposed~\cite{srt}, and with our improved architecture (\oursrt, \ourupsrt).
    Despite requiring no poses, \rust\ matches the performance of SRT and \ourupsrt\ while strongly outperforming all methods when target pose is noisy \tpbad.
    \textbf{Right:}
    Model ablations, see \cref{sec:exp:ablations}.
}
\label{tab:psnr}
\end{table}

\section{Experiments}
\label{sec:exp}

We begin our experiments on the MultiShapeNet (MSN) dataset (version from~\cite{osrt}),
a very challenging test bed for 3D-centric neural scene representation learning~\cite{srt}.
It consists of synthetically generated 3D scenes~\cite{greff2021kubric} that contain \mbox{16-32} ShapeNet objects~\cite{chang2015shapenet} scattered in random orientations, and with photo-realistic backgrounds.
Importantly, the set of ShapeNet objects is split for the training and test datasets, meaning that all objects encountered in the test scenes are not only in novel arrangements and orientations, but the model has never seen them in the training dataset.
Camera positions are sampled from a half-sphere with varying distances to the scene.
In all experiments, we follow prior work~\cite{srt} by using 5 input views, and all quantitative metrics are computed on the right halves of the target views, since the left halves are used for pose estimation in \rust.

\begin{figure}[t]
\centering
\includegraphics[trim={0 7 0 7},clip,width=1.0\columnwidth]{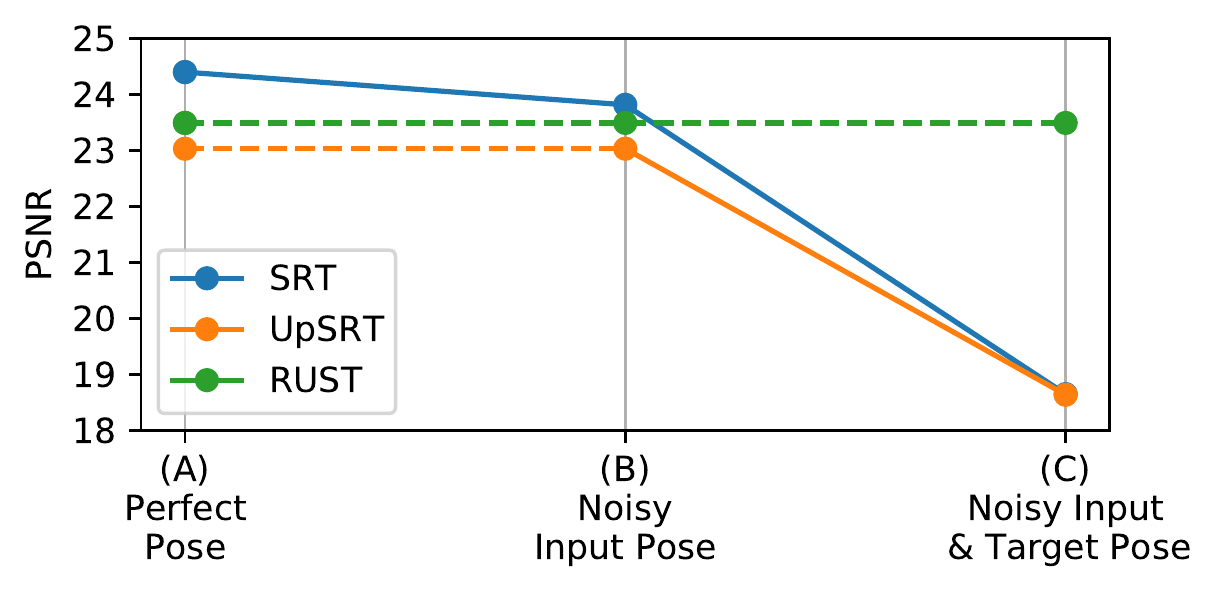}
\caption{
\textbf{Robustness to camera noise} --
\citet{srt} evaluate SRT and UpSRT on (A) perfect pose, and (B) noisy \emph{input} pose.
In the more realistic setting (C) where \emph{input \& target} pose is noisy, both methods fail as they rely on accurate target camera pose for training.
\rust\ needs no pose, so its performance is constant.
} 
\label{fig:camnoise}
\end{figure}

\subsection{Novel view synthesis}
\label{sec:exp:nvs}

The core contribution of \rust\ is its ability to model 3D scenes without posed imagery.
While camera poses in synthetic environments such as MSN are perfectly accurate, practical applications must often rely on noisy sensor data or inaccurate estimated poses.
We therefore begin our investigations by measuring the quality of synthesized novel views in new scenes under various assumptions on the accuracy and availability of camera pose information during training.
To simulate the real world, we follow prior work~\cite{srt, Lin21iccv_BARF} and perturb the training camera poses with additive noise for a relatively mild amount of $\sigma$\,$=$\,$0.1$.
We visualize the effect of this noise in \cref{fig:app:camnoise}
(appendix) to provide context for its scale.
This leads to three possible settings for both input poses and target poses: perfect (\ipgood, \tpgood), noisy (\ipbad, \tpbad) and lack of (\ipnone, \tpnone) input and target cameras poses, respectively.
Note that for the posed baselines, we always use perfect target poses for evaluation, \ie the target poses are only perturbed during training.

\cref{tab:psnr} (left) shows our quantitative evaluation on the MSN dataset.
We first compare SRT as proposed~\cite{srt} with our improved architecture \oursrt, both using perfect pose.
We observe that our modifications to the architecture lead to significant improvements in PSNR (+1.09\,db).
Continuing with the improved architecture, perturbing the input views leads to a loss in 0.59\;db (\oursrt, \ipbad, \tpgood) while the lack thereof leads to a drop of 1.37\;db (\ourupsrt, \ipnone, \tpgood).
\def\msvpad{-1mm}

\def\msnsize{0.137\linewidth}
\newcommand\mspica[1]{
    \includegraphics[width=\msnsize]{imgs/qualitative/a/#1}
}
\newcommand\mspicb[1]{
    \includegraphics[width=\msnsize]{imgs/qualitative/b/#1}
}

\begin{figure*}[t]
    \centering 
    \setlength{\tabcolsep}{0.1mm}
    \def\arraystretch{2}
    \begin{tabular}
    {ccccccc}
    \multicolumn{1}{c}{Target} &
    \multicolumn{1}{c}{SRT~\cite{srt}} &
    \multicolumn{2}{c}{\oursrt} &
    \multicolumn{2}{c}{\ourupsrt} &
    \multicolumn{1}{c}{\rust}
    \\[-2mm]
    \cmidrule(lr){1-1}
    \cmidrule(lr){2-2}
    \cmidrule(lr){3-4}
    \cmidrule(lr){5-6}
    \cmidrule(lr){7-7}
    \\[-12mm]
    \multicolumn{1}{c}{} &
    \multicolumn{1}{c}{\ipgood,\; \tpgood} &
    \multicolumn{1}{c}{\ipgood,\; \tpgood} &
    \multicolumn{1}{c}{\ipbad,\; \tpgood} &
    \multicolumn{1}{c}{\ipnone,\; \tpgood} &
    \multicolumn{1}{c}{\ipnone,\; \tpbad} &
    \multicolumn{1}{c}{\ipnone,\; \tpnone}
    \\[0mm]
    \mspica{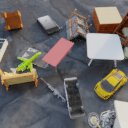} &
    \mspica{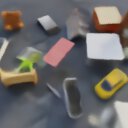} &
    \mspica{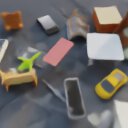} &
    \mspica{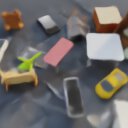} &
    \mspica{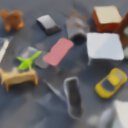} &
    \mspica{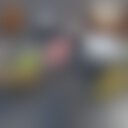} &
    \mspica{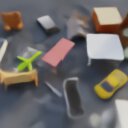}
    \\ [\msvpad]
    \mspicb{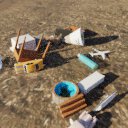} &
    \mspicb{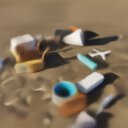} &
    \mspicb{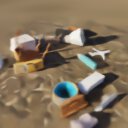} &
    \mspicb{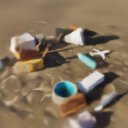} &
    \mspicb{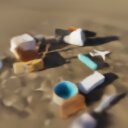} &
    \mspicb{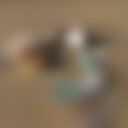} &
    \mspicb{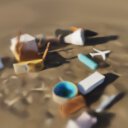}
    \\ [-3mm]
    \end{tabular}
    \caption{
    \textbf{Qualitative results on MSN} --
    We compare \rust to SRT~\cite{srt} as proposed, and with the respective models using our improved architecture (\oursrt, \ourupsrt).
    Different models support perfect (\ipgood, \tpgood), noisy (\ipbad, \tpbad), or a lack of (\ipnone, \tpnone) input and target poses, respectively.
    \oursrt\ shows a mild drop in quality for perturbed input poses and achieves a similar reconstruction quality as \ourupsrt\ with perfect target poses.
    In the realistic setting where the target poses are also perturbed (\tpbad), both methods fail entirely to render sharp images (only \ourupsrt\ shown here, see appendix for \oursrt).
    \rust, without any camera poses, produces similar-quality renders as the baselines with perfect pose information, and even visibly outperforms the original SRT model~\cite{srt} with perfect pose information.
    }
\label{fig:qualitative}
\end{figure*}

Crucially, the most realistic setting where \emph{all} poses are equally noisy leads to a dramatic decline in performance for prior work (\oursrt\ and \ourupsrt, \tpbad).
\rust\ meanwhile outperforms the baselines by +4.84\;db in this setting without any poses, while matching the performance of SRT~\cite{srt} with perfect pose.
We highlight the difference between only perturbing input poses (following prior work~\cite{srt}) or all camera poses (more realistic setting) in \cref{fig:camnoise}.
It is evident from the plot that \rust\ is the only model that is applicable when camera poses are noisy or even unavailable.

\cref{fig:qualitative} shows qualitative results for a selection of these models.
Notably, \ourupsrt\ produces extremely blurry views of the scene when the target poses are mildly inaccurate, while \rust\ produces similar-looking results as \ourupsrt\ with perfect target poses.
Further qualitative results including videos are provided in the supplementary material.

For the remainder of the experimental section, we only compare \rust\ to the stronger \oursrt\ and \ourupsrt\ baselines and refer to these as SRT and UpSRT for ease of notation.

\vspace{-2mm}
\subsubsection{Ablations}
\label{sec:exp:ablations}
\vspace{-1mm}

We investigate the effect of a selection of our design choices for \rust.
The results are summarized in \cref{tab:psnr} (right).
Metrics are computed only on the right half of each target image, since the left half is used for pose estimation by \rust.
While we see a lot of evidence that the model is encoding a form of camera pose in the latent space (see \cref{sec:exp:lpi}), it is still possible that the model uses parts of the latent pose feature to encode \emph{content} information about the target view.
We therefore now evaluate the exact same \rust\ model (with identical weights), but now use the right halves of the target images in the Pose Estimator.
We observe that this scheme only outperforms the left-half encoding scheme by +0.39\,db, thereby showing that the latent pose primarily serves as a proxy for the camera position rather than directly informing the model about the target view content.

In our default model, we allow gradients to flow from the latent pose back into the encoder, though they are effectively scaled down by a factor of 0.2 (see \cref{sec:method:training}).
Cutting the gradients fully implies that the the Pose Estimator cannot directly affect the encoder anymore through the SLSR during training.
This variant leads to a drop of 0.33\,db in PSNR.

\newcommand\msntrgraphics[1]{\includegraphics[width=0.157\linewidth]{imgs/msntr2/#1}}

\begin{figure*}[t]
    \centering
    \begin{minipage}{0.25\linewidth}
        \includegraphics[width=\linewidth]{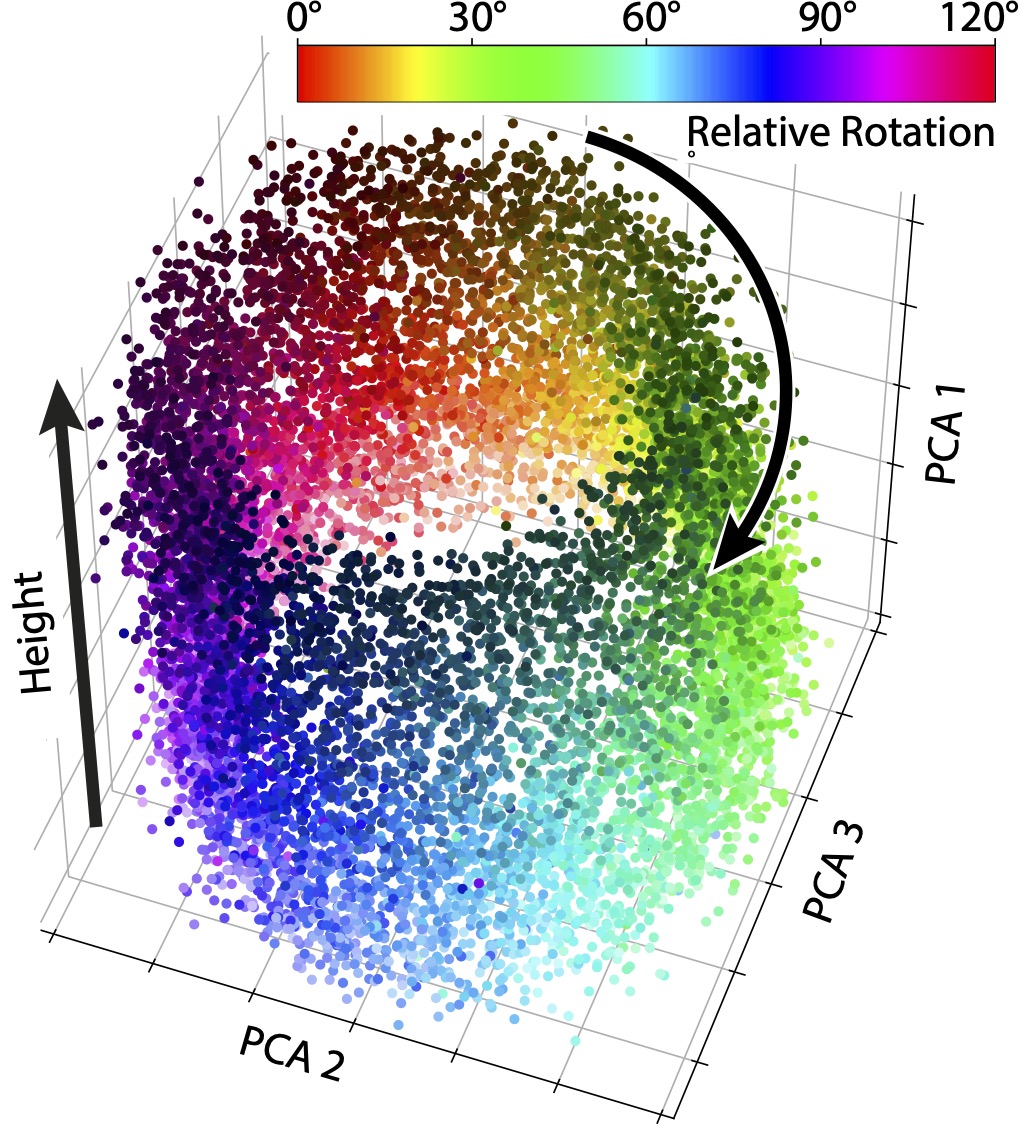}
    \end{minipage}
    \begin{minipage}{.735\textwidth}
        \centering 
        \setlength{\tabcolsep}{0.2mm}
        \def\arraystretch{2}
        \begin{tabular}{cc@{\hskip 1mm}cc@{\hskip 1mm}cc}
            \multicolumn{2}{c}{Height} &
            \multicolumn{2}{c}{Rotation} &
            \multicolumn{2}{c}{Distance}
            \\[-2mm]
            \cmidrule(lr){1-2}
            \cmidrule(lr){3-4}
            \cmidrule(lr){5-6}
            \\[-8mm]
            \msntrgraphics{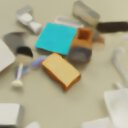} &
            \msntrgraphics{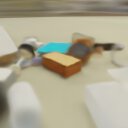} &
            \msntrgraphics{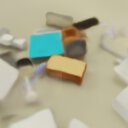} &
            \msntrgraphics{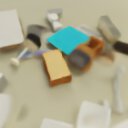} &
            \msntrgraphics{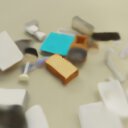} &
            \msntrgraphics{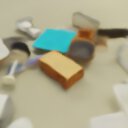}
            \\ [-2mm]
            \msntrgraphics{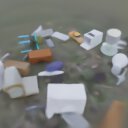} &
            \msntrgraphics{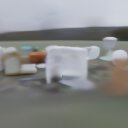} &
            \msntrgraphics{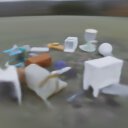} &
            \msntrgraphics{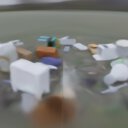} &
            \msntrgraphics{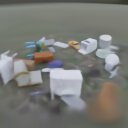} &
            \msntrgraphics{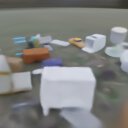} \\
        \end{tabular}
    \end{minipage}
    \caption{
    \textbf{Latent pose investigations on MSN} --
    \textbf{Left:}
    The first three PCA components of the latent poses across many scenes roughly constitute a cylinder.
    Points are coloured such that intensity represents camera height and hue represents their rotation around the scene's z-axis relative to the first input camera, modulo $120^\circ$.
    That means a single rotation around the scene maps to three rotations around this circle, with the $0^\circ$ position mapping to the first input camera.
    We found that components 5 and 6 roughly capture one full 360 degree rotation, resolving the ambiguity.
    We elaborate on this in \cref{app:sec:msn_pose_investigations}.
    \textbf{Right:}
    Intuitive camera movement induced by traversals in PCA space.
    }
    \label{fig:msn:traversals}
\end{figure*}

As described in \cref{sec:method:poseestimator}, the Pose Estimator cross-attends into parts of the SLSR to allow it to anchor the target poses relative to the input views.
Removing this cross-attention module makes it much harder for the model to estimate the target pose and pass it to the decoder.
The performance therefore drops by a significant 2.66\,db in PSNR.
Removing the self-attention between the target image tokens from the Pose Estimator module has a less dramatic effect
of 0.52\,db.

Finally, we investigate different choices for the size of the latent pose feature \latentpose.
We found empirically that smaller sizes such as 3 dimensions would lead to significantly worse results (-3.09\,db in PSNR).
This is likely only a result of worse training dynamics, as the pose space in the MSN dataset can in theory be fully described with 3 degrees of freedom.
Using significantly larger latent pose sizes (64, 768) on the other hand leads to a less controllable latent pose space, while rendering quality remains comparable.

\subsection{Latent pose investigations}
\label{sec:exp:lpi}

In order to analyze the structure of the learned latent pose space, we use principal component analysis (PCA).
Specifically, we collect latent poses \latentpose\ for three target views per scene from 4\:\!k test scenes of the MSN dataset and inspect the major PCA components of the resulting 8-dimensional pose distribution.
We visualize the first three PCA components in \cref{fig:msn:traversals} (left) and color-code the points such that hue encodes rotation relative to the first input view's camera and intensity encodes camera height.
We find that the points form a cylinder whose axis is aligned with the first principal component.
This axis correlates strongly with camera height with a Pearson's $r$\,$=$\,$0.94$.
The qualitative effect of moving along this dimension is shown in~\cref{fig:msn:traversals} (\emph{Height}).
Similarly, rotating around the cylinder causes the camera to rotate around the scene in 3D as shown in~\cref{fig:msn:traversals} (\emph{Rotation}), although only a third of the rotation is completed by a full traversal along the cylinder.
Note that rotations in the latent space are encoded relative to the position of the first input camera, likely because the Pose Estimator only consumes the SLSR tokens $\tilde{\slsr}$ corresponding to the first input view.

The cylinder-shaped latent pose distribution is sensible, considering that the camera poses in the MSN dataset are distributed along a dome and always point at the scene origin.
At the pole of this dome is hence a discontinuity as the camera flips when crossing that point.
The cylinder is therefore akin to this dome that is opened up at its pole to eliminate the discontinuity.
The remaining PCA components are visualized in \cref{app:sec:msn_pose_investigations}.
The fourth component captures the distance of the camera from the scene center ($r$\,$=$\,$0.94$).
We show traversals along this axis in \cref{fig:msn:traversals} (\emph{Distance}).
We further find that components 5 and 6, capture a 360 degree \emph{absolute} rotation in scene coordinates.
This surprising finding is further investigated in \cref{app:sec:msn_pose_investigations}.

It is notable that the model has learned this meaningful structure in the latent pose space without any form of camera pose supervision.
Furthermore, this shows that test-time camera control is feasible directly in the latent space.

\begin{table}[t]
\centering
\vspace{2mm}
\begin{tabular}{lcccc}
\toprule
Method & \#\,Views & MSE & $R^2$ ($\%$) & Success ($\%$)
\\

\midrule
\rust\ EPE & 7 & 0.08 & 99.9 & [100]  \\ %

\midrule
COLMAP & 10  & 0.00   & 100.0 &  4.2  \\
COLMAP & 80  & 0.07   & 99.7  & 29.5  \\
COLMAP & 160 & 0.38   & 99.1  & 58.9  \\

\midrule
GNeRF & 12 & 29.39   & 46.7  & [100]  \\
GNeRF & 150 & 9.24   & 83.1  & [100]  \\
GNeRF-FG & 150 & 4.05   & 92.7  & [100]  \\

\bottomrule
\end{tabular}
\caption{
    \textbf{Explicit pose estimation on MSN} --
    RUST EPE recovers relative camera poses nearly perfectly from the SLSR (5 input views) and the pair of latent target poses.
    COLMAP~\cite{colmap} requires a much larger number of images, and still has a significantly lower success rate for registration.
    Similarly, GNeRF~\cite{Meng21iccv_GNeRF} requires many views of the scene, and fails to estimate accurate poses even when the background pixels are removed from the data (GNeRF-FG).
}
\label{tab:epe}
\end{table}

\vspace{-2mm}
\subsubsection{Explicit pose estimation}
\label{sec:exp:epe}
\vspace{-1mm}

To validate to what extent information about the camera pose is retained in the small latent pose $\tilde{p}$ of \rust, we perform an explicit pose estimation (EPE) experiment.
As there is no canonical frame of reference, we choose to predict \emph{relative} poses between two separate target views.
Since all cameras on the MSN dataset are pointed at the center of the scene, we only estimate the relative camera \textit{position}.

Our EPE module is trained on top of a (frozen) pre-trained \rust\ model.
It takes the SLSR $\slsr$ and the latent pose features $\tilde{p}_1$ and $\tilde{p}_2$ for the two target views and follows the design of the \rust\ decoder: the concatenation of the two latent poses acts as the query for a transformer which cross-attends into the SLSR $\slsr$.
We found that this cross-attention step is necessary as the latent pose vectors $\tilde{p}$ partially carry pose information that is relative to the SLSR (\eg, rotation, see \cref{sec:exp:lpi}).
Finally, the result is passed through an MLP which is tasked to predict the explicit relative pose between the target views.
We train the model using the L2 loss.

Quantitative results in terms of mean-squared error (MSE) and $R^2$ scores over 95 MSN test scenes are shown in \cref{tab:epe}.
EPE recovers relative camera positions nearly perfectly on novel test scenes ($R^2$\,$=$\,$99.9\%$) using only the SLSR (derived from 5 input views) and the pair of 
latent pose vectors.

\vspace{-1mm}
\paragraph{Photogrammetry}
To demonstrate the difficulty of the task, we apply COLMAP~\cite{colmap} to 95 newly generated MSN test scenes with up to 160 views (otherwise using the same parameters as the MSN test set).
COLMAP uses correspondences between detected keypoints to estimate camera poses.
While RUST EPE successfully recovers camera positions from just 7 total views, COLMAP struggles when only given access to a small number of views (\eg, $< 10$) and needs up to 160 views to successfully register most views and estimate their camera poses.
As COLMAP may pose only a subset of the provided images, we report metrics for a predetermined, arbitrarily chosen camera pair per scene.
If COLMAP fails to pose either of these cameras, we consider pose estimation to have failed and exclude this camera pair from evaluation.
We find an optimal rigid transformation that maps estimated camera positions to ground-truth positions before evaluating MSE and $R^2$ to be able to compare with RUST EPE.

\vspace{-2mm}
\paragraph{NeRF-based methods}
\def\gnerfsize{0.19\columnwidth}
\newcommand\gnerfpic[1]{
    \includegraphics[
        width=\gnerfsize
    ]{imgs/gnerf2/#1}
}

\begin{figure}[t]
    \centering 
    \setlength{\tabcolsep}{0.0mm}
    \def\arraystretch{1}
    \begin{tabular}
    {ccccc}
    \small Target &
    \small 150 w/ pose &
    \small 12 w/ pose &
    \small 150 &
    \small 150 FG
    \\
    \gnerfpic{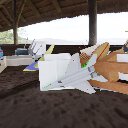} &
    \gnerfpic{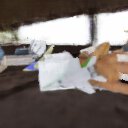} &
    \gnerfpic{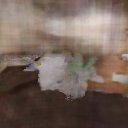} &
    \gnerfpic{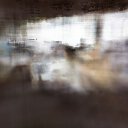} &
    \gnerfpic{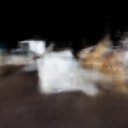}
    \\
    \small PSNR &
    \small 21.49 &
    \small 13.41 &
    \small 14.42 &
    \small 15.71
    \end{tabular}
    \caption{
    \textbf{Qualitative results for GNeRF~\cite{Meng21iccv_GNeRF}} --
        When trained with many posed views (150), results have high fidelity.
        Reducing the number of images to 12 leads to worse reconstruction and strong artifacts.
        When no pose is available, the model produces worse quality.
        Removing the backgrounds from the training dataset helps improve reconstruction accuracy.
        PSNR computed on 50 views.
    \vspace{-2mm}
    }
\label{fig:gnerf}
\end{figure}

We evaluate GNeRF~\cite{Meng21iccv_GNeRF} as a strong representative for unsupervised NeRF-based pose estimation approaches on a single MSN scene using the implementation provided by the authors.
GNeRF assumes knowledge of the camera intrinsics and that all cameras are pointed at the origin, \ie only the 3-dimensional camera positions are estimated.
Further, a prior for the camera pose initializations is given to the model.
Using only 12 views, GNeRF fails to capture the scene.
When using 150 views to train GNeRF, it still performs significantly worse than \rust\ EPE on explicit pose estimtation, demonstrating the benefit of learning a generalizable pose representation \textit{across} scenes, see \cref{tab:epe}.
We show qualitative and quantitative results for novel view synthesis in \cref{fig:gnerf}.
Training on perfect poses significantly improves NVS performance for GNeRF, showing that even with 150 views, GNeRF fails to accurately register the camera poses.
As GNeRF expects the scene to be contained within a given bounding box, we also try training it only on the foreground (FG) objects in the scene, dropping all background pixels during optimization.
We empirically find that this only leads to slightly improved pose accuracy.

\subsection{Application to real-world dataset}
\label{sec:exp:sv}
\begin{figure}[t]
\centering
\includegraphics[width=\columnwidth]{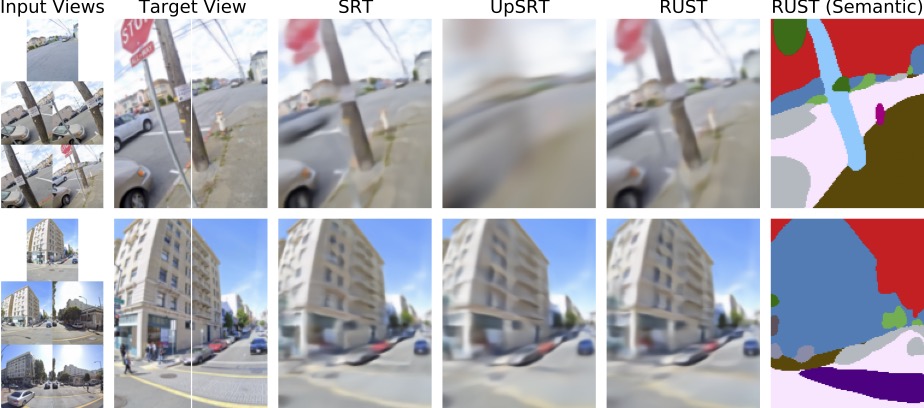}
\caption{
    \textbf{Qualitative results on SV} --
    Comparison of \rust\ with prior work using accurate camera pose.
    \rust\ outperforms our improved UpSRT variant, while producing similar quality as the fully posed improved SRT model.
    We further train a dense semantic segmentation decoder on top of the frozen \rust\ scene representation, showing that it retains semantic information about the scene.
}
\label{fig:sv:qualitative}
\end{figure}

To answer the question whether \rust\ can learn to represent the 3D structure of complex real-world scenes, we apply it to the Street View (SV) dataset.
We have received access to SV through private communication with the authors~\cite{GoogleStreetView}. 
It contains 5\:\!M dynamic street scenes of San Francisco with moving objects, changes in exposure and white balance, and highly challenging camera positions, often with minimal visual overlap.
Following \citet{srt}, we train and test the model using 5 randomly selected input views for each scene.

\def\svsepsize{0.5mm}
\def\svsize{0.145\linewidth}
\newcommand\svpic[2]{
    \includegraphics[width=\svsize]{imgs/sv_interpolation/#1_#2}
}

\begin{figure*}
\begin{minipage}{.37\textwidth}
    \centering
    \vspace{7mm}
    \includegraphics[width=0.9\textwidth]{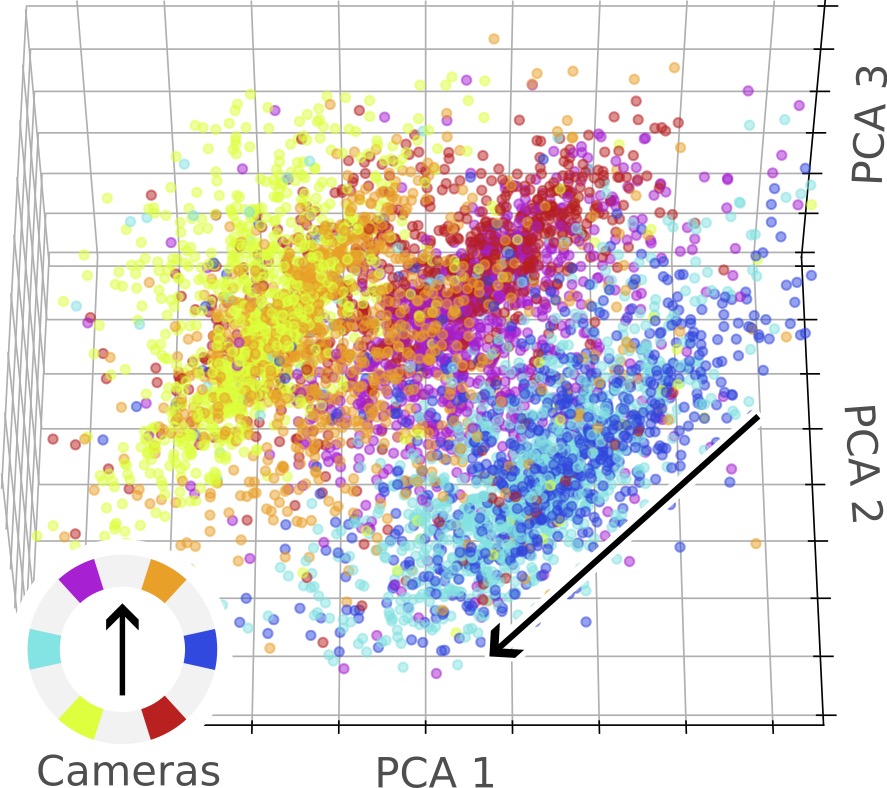}
\end{minipage}%
\begin{minipage}{.61\textwidth}
    \centering 
    \setlength{\tabcolsep}{0.0mm}
    \def\arraystretch{2}
    \begin{tabular}{rlrlrl}
    \multicolumn{2}{c}{Backward\;/\;Forward} &
    \multicolumn{2}{c}{Up\;/\;Down} &
    \multicolumn{2}{c}{Tilt Right\;/\;Left}
    \\[-2mm]
    \cmidrule(lr){1-2}
    \cmidrule(lr){3-4}
    \cmidrule(lr){5-6}
    \\[-8mm]
    \hspace{\svsepsize}\svpic{6_4}{bwd} &
    \svpic{6_4}{fwd}\hspace{\svsepsize} &
    \hspace{\svsepsize}\svpic{6_4}{down} &
    \svpic{6_4}{up}\hspace{\svsepsize} &
    \hspace{\svsepsize}\svpic{6_4}{unroll} &
    \svpic{6_4}{roll}\hspace{\svsepsize}
    \\ [-1mm]
    \svpic{33_2}{bwd} &
    \svpic{33_2}{fwd} &
    \svpic{33_2}{down} &
    \svpic{33_2}{up} &
    \svpic{33_2}{unroll} &
    \svpic{33_2}{roll} 
    \\
    \end{tabular}
\end{minipage}
\caption{
    \textbf{Latent pose investigations on SV} --
    \textbf{Left:} Scatterplot showing the first three PCA components of the latent pose \latentpose\ distribution for 12\:\!k target views from 4\:\!k scenes.
    The color of each point indicates by which of the 6 cameras the image was taken, as shown on the color wheel.
    The first principal axis of these ellipsoids, marked by the long arrow, corresponds to movement along the street.
    \textbf{Right:} We found that specific linear directions in the latent space correspond to meaningful camera motions.
    Videos are provided in the supplementary material.
}
\label{fig:sv:interpolation}
\label{fig:sv:scatter}
\end{figure*}

\vspace{-1mm}
\paragraph{Novel view synthesis}
We first evaluate the novel view synthesis performance of \rust, and again find that its performance of 22.50\,db in PSNR is comparable to that of SRT (22.72) and that it falls in-between that of our improved SRT (23.63) and UpSRT (21.25).
Qualitatively, we find that novel views generated by \rust\ correctly capture the 3D structure of the scenes and, remarkably, the model seems to be more robust than UpSRT, see \cref{fig:sv:qualitative}.
This happens especially often when the reference view is far away from the target view.

\vspace{-1mm}
\paragraph{Latent pose space}
In SV scenes, the camera positions mainly vary along the single forward-dimension of the street while horizontal rotations mainly move in discrete steps of $60^{\circ}$ due to the spatial configuration of the six fixed cameras.
This distribution of camera poses is very different from MSN, and we should thus expect the learned pose space to differ significantly as well.
Similar to before, we compute a PCA decomposition on target views of 4\:\!k test-scenes and visualize the first three components in \cref{fig:sv:scatter}.
Surprisingly, we find only three (rather than six) ellipsoid clusters that each correspond to an opposing pair of cameras.
This suggests that the model has learned to represent rotation in an absolute (scene-centric) coordinate frame with the direction of the street as a reference, and that opposing cameras are less well-distinguished due to the streets' approximate mirror symmetries, see \cref{app:fig:confusion} (appendix) for more details.
We further find that movement along the street corresponds well to the long axis of these ellipsoids, which depending on the camera can correspond to either a forward/backward or a sideways motion in image space.

\vspace{-3mm}

\paragraph{Traversals}
In \cref{fig:sv:interpolation}, we show linear latent pose traversals that correspond to movement along the street, pitch and tilt of the camera. 
We find that these dimensions are mapped into the pose space predictably, contiguously and smoothly.
The correct parallax effect observed in the forward movement in particular demonstrates that the model has correctly estimated the depth of the scene.
These results indicate that RUST can learn to capture the 3D structure of complex real-world scenes from a handful of images, without access to any pose information.
The learned latent pose space covers the training distribution of camera poses and enables traversals and novel view synthesis within distribution.

\section{Limitations}
\label{sec:limitations}

\rust\ shares most strengths and limitations with other latent methods, especially with the SRT model~\cite{srt} that it is based on.
Resolution and quality of synthesized views do not reach the performance of NeRF~\cite{Mildenhall20eccv_nerf}, though it is important to note that NeRF does not yield latent scene representations, and it generally requires a large number of similar views of the scene.
As has been demonstrated~\cite{srt}, even specialized NeRF variants for sparse input view settings~\cite{Jain21iccv_DietNeRF} fail to produce meaningful results on the challenging SV dataset, despite having access to accurate pose information.

As shown in \cref{sec:exp:lpi,sec:exp:sv}, \rust\ learns to effectively model the camera pose distribution of the datasets it is trained on.
While this is desirable in many applications, it can limit generalization of the model to new poses that lie outside of the original distribution.
For example, we may not find latent poses for which RUST would render views that do not point at the center of the scene in MSN, or represent movement orthogonal to the street in SV.
It is notable that all learned methods suffer from such limitations to varying degrees, including light field parametrizations~\cite{srt} and even volumetric models, especially when given few input images~\cite{niemeyer2022regnerf}.
Nonetheless, more explicit methods tend to generalize better to out-of-distribution poses due to hard-coded assumptions such as explicit camera parametrizations or the volumetric rendering equation.
Finally, we empirically found \rust\ to show higher variance in terms of PSNR compared to SRT.
Over independent training runs with 3 random seeds, we observed a standard error of 0.19.

\section{Conclusion}
\label{sec:conclusion}
We propose \rust, a novel method for latent neural scene representation learning that can be trained without any explicit pose information.
\rust\ matches the quality of prior methods that require pose information while scaling to very complex real-world scenes.
We further demonstrate that the model learns meaningful latent pose spaces that afford smooth 3D motion within the distribution of camera poses in the training data.
We believe that this method is a major milestone towards applying implicit neural scene representations to large uncurated datasets.

\clearpage
{\small
\bibliographystyle{plainnat}
\bibliography{main}
}

\clearpage
\appendix
\section{Appendix}
\label{sec:app}

\paragraph{Acknowledgements}
We thank Urs Bergmann for helpful feedback on the manuscript and Henning Meyer and Thomas Unterthiner for fruitful discussions.

\subsection{Model details [\cref{sec:method}]}
\label{app:sec:model_details}

Unless stated otherwise, model architecture and hyper parameters follow the original SRT~\cite{srt}.
Our improved architecture makes the following changes that are partially inspired from prior work \cite{osrt}.
In the encoder CNN, we drop the last block, leading to smaller 8$\times$8 patches compared to the 16$\times$16 patches as in SRT~\cite{srt}.
This reduces the parameter count of the CNN and makes it faster, while the subsequent encoder transformer needs to process 4$\times$ more tokens.
However, the number of parameters in the encoder transformer is not affected by this change.
In the encoder transformer, we use 5 layers instead of 10.
Instead of using post-normalization, we use pre-normalization.
We further use full self-attention instead of cross-attending into the output of the CNN.
The initial queries in the decoder pass through a linear layer that maps them to 256 dimensions.

All models (SRT~\cite{srt}, \oursrt, \ourupsrt, \rust) were always trained for 3\,M steps with batch size 256 on MSN and 128 on SV.
Similar to SRT~\cite{srt}, we note that \rust\ is not fully converged yet at this point, though improvements are comparably small when trained further.

\vspace{-1mm}
\paragraph{Pose Estimator}
The CNN in the Pose Estimator follows very similar structure as the CNN in the encoder, with two exceptions:
we use 4 CNN blocks as in SRT~\cite{srt} (\ie, 16$\times$16 patches), and we do not add image id embeddings, as each target view is processed and rendered fully independently.
The transformer follows the same structure as the encoder transformer: we use 768 features for the attention process and 1536 features in the MLPs and pre-normalization.

\vspace{-1mm}
\paragraph{Appearance encoder}
For all models on SV, we additionally include the appearance encoder from the original SRT~\cite{srt}.
Empirically, we have seen that leaving out the appearance encoder in \rust\ leads to the Position Encoder capturing variations in exposure and while balance in \latentpose, making the appearance encoder obsolete -- but for better comparability with the baselines, we opted to include it for our model as well.
This design choice also has the positive side effect that the latent pose space is disentangled from variations in appearance.

\begin{figure}[t]
    \centering 
    \includegraphics[width=0.8\columnwidth]{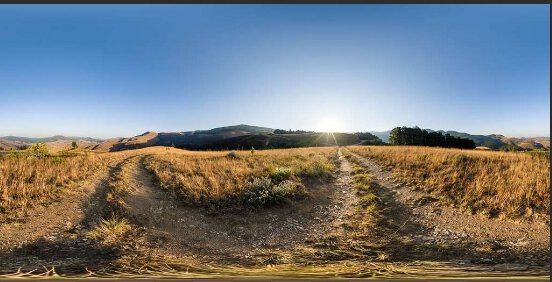}
    \includegraphics[width=0.8\columnwidth]{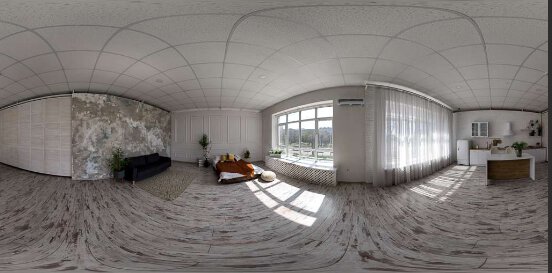}
    \includegraphics[width=0.8\columnwidth]{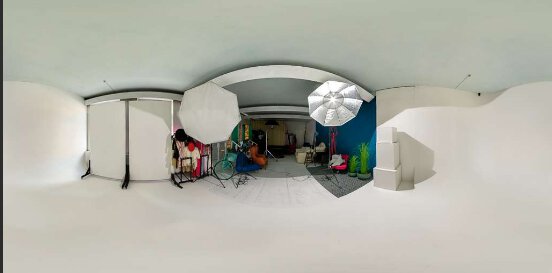}
    \includegraphics[width=0.8\columnwidth]{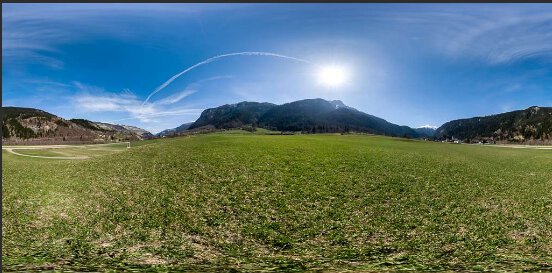}
    \includegraphics[width=0.8\columnwidth]{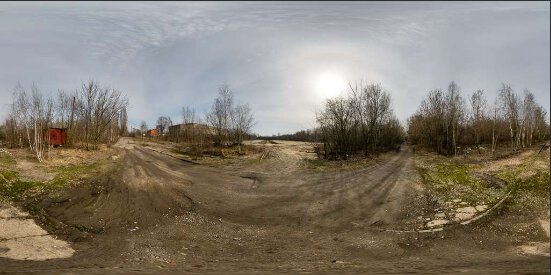}
    \caption{
    \textbf{Example backgrounds from HDRi Haven} -- Note that the main light source is always slightly to the right of the center.
    }
\label{fig:app:hdrihaven}
\end{figure}

\subsection{Latent pose investigations on MSN [\cref{sec:exp:lpi}]}
\label{app:sec:msn_pose_investigations}
In \cref{fig:app:msn:pca56} we visualize the latent pose dis\-tri\-bu\-tion along PCA components 4, 5, and 6, and  demonstrate full $360^{\circ}$ camera rotations around the scene by interpolating poses.
The scatter plot of poses on the left side of \cref{fig:app:msn:pca56} shows camera poses colored by their absolute ground-truth rotation. 
The fact that the learned poses map so well onto the ground-truth coordinate frame of the generated scenes is surprising, since the model at no point has access to those coordinates, and the scene layout is mostly rotation symmetric.
It turns out that the reason for this alignment has to do with the $360^{\circ}$ HDR images from \href{https://hdri-haven.com/}{hdri-haven.com} used as backgrounds and for global illumination in MSN.
These images are aligned such that the strongest light source (often the sun, but sometimes, windows or lamps as well) is always in roughly the same direction, see \cref{fig:app:hdrihaven}. 
The background images are not randomly rotated between scenes, so the main direction of light correlates strongly with the ground-truth coordinate frame. 
And, as \cref{fig:app:msn:pca56} shows, it thus also correlates strongly with the coordinate frame learned by RUST.

\subsection{Explicit Pose Estimation [\cref{sec:exp:epe}]}
\label{app:sec:epe}

The explicit pose estimation (EPE) module for RUST follows the architecture of the RUST decoder: we take the RUST latent pose for a target view and a reference view of a scene, concatenate both vectors along the channel dimension, and project the result to a 768-dim vector using a learnable Dense layer. The result is used as a query for a transformer that cross-attends into the SLSR with 2 layers, 12 heads, query/key/value projection dimension of 768, and an MLP hidden dimension of 1536 for the feed-forward layer, using pre-normalization and trained without dropout.

\begin{table}[t]
\centering
\vspace{2mm}
\begin{tabular}{cccc}
\toprule
\#\,Views 
    & MSE 
    & $R^2$ ($\%$) 
    & Success ($\%$)
\\
\midrule
10  & 0.00 & 100.0 & 4.2  \\
20  & 0.01 & 99.8  & 5.3  \\
40  & 0.01 & 99.9  & 16.8 \\
80  & 0.07 & 99.7  & 29.5 \\
120 & 0.56 & 97.8  & 43.2 \\
160 & 0.38 & 99.1  & 58.9 \\
200 & 0.28 & 99.3  & 66.3 \\
\bottomrule
\end{tabular}
\caption{
    \textbf{Explicit Pose Estimation with COLMAP} --
    COLMAP requires between 120 and 160 images per scene to successfully pose the designated image pair more than half of the time.
    The perfect $R^2$ score in the first row is an artifact of the low success rate and our alignment procedure for evaluation.
}
\label{tab:app_epe_colmap}
\end{table}

\begin{figure}[t]
\centering
\includegraphics[trim={2cm 15.5cm 2cm 18mm},clip,width=\columnwidth]{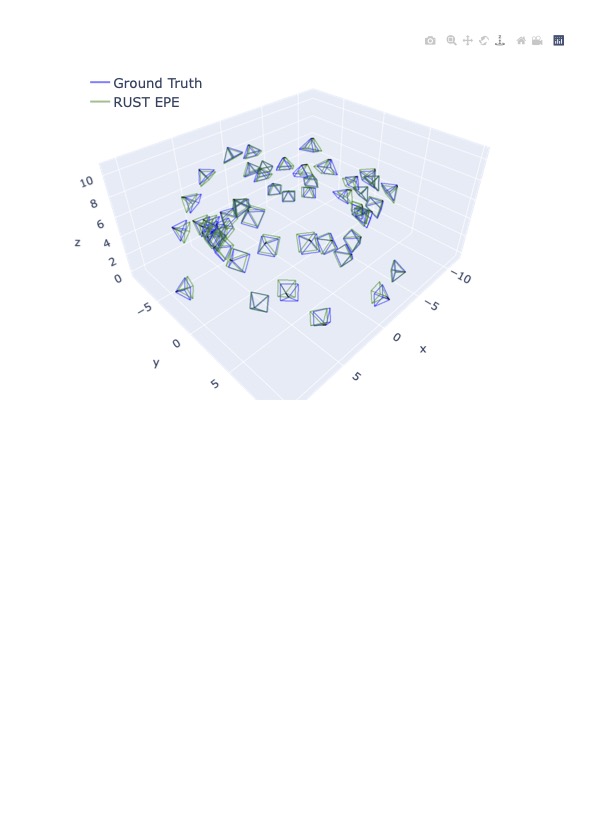}
\caption{
\textbf{RUST EPE} --
Camera positions recovered by RUST EPE (oriented towards zero-center solely for visualization).
RUST EPE estimates (green) are obtained relative to randomly chosen reference views, and very closely match the ground truth (blue).
}
\label{fig:app:rust_epe}
\end{figure}

The result is finally used to predict the 3-dim difference between the target and reference camera origins using an MLP with a single 768-dim hidden layer. We train the EPE module on top of a frozen, pre-trained RUST model using a mean-squared error loss between prediction and ground-truth camera origin difference (target origin $-$ reference origin). The pre-trained RUST model was trained on MSN for 3M steps (i.e.~the same model that is used in other experiments). We train RUST EPE for 1M steps on the MSN training set using the Adam optimizer with otherwise the same hyperparameters as RUST. We evaluate RUST EPE on the first 95 (unseen) test set scenes. 
Estimated camera views (relative to arbitrarily chosen reference views) are shown in \cref{fig:app:rust_epe}.

\paragraph{COLMAP}
As a baseline, we compare RUST EPE to COLMAP~\cite{colmap}, a general purpose Structure-from-Motion (SfM) pipeline for estimating camera parameters.
Unlike RUST, COLMAP requires high visual overlap between image pairs to identify ``feature matches'' -- pixel locations corresponding to the same point in 3D space.
In preliminary experiments, we found that COLMAP is unable to reconstruct scenes with fewer than 10 images.
We thus generate an evaluation set of 95 scenes, each with 200 frames, from the same distribution as the MSN dataset.

\begin{figure}[t]
\centering
\includegraphics[trim={2cm 15.5cm 2cm 2cm},clip,width=\columnwidth]{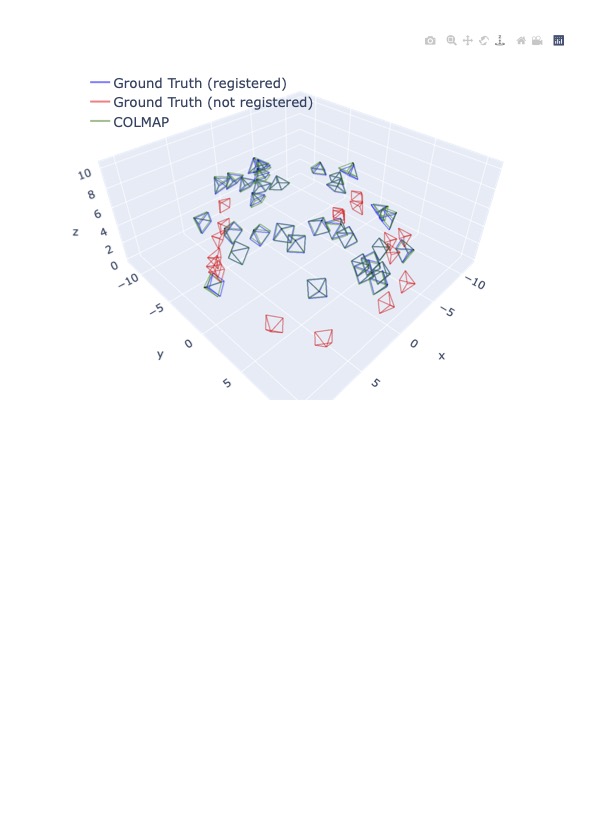}
\caption{
\textbf{COLMAP} --
Camera positions recovered by COLMAP (oriented towards zero-center solely for visualization).
For images with estimated camera pose, COLMAP's estimates (green) are extremely close to the ground truth (blue).
COLMAP failed to pose ground truth cameras in red.
}
\label{fig:app_colmap}
\end{figure}

We evaluate COLMAP as follows.
We begin by running COLMAP's SfM pipeline using the first 10 / \ldots\ / 200
images of each scene, sorted lexicographically.
As the coordinate system of the estimated camera origins does not align with that of the ground-truth cameras, we find the optimal rigid transformation (rotation + translation + scale) that minimizes the squared error between camera origin pairs (predicted and ground-truth).
To evaluate the relative pose, we select the camera origins of two fixed, arbitrary images (\texttt{camera\_1} and \texttt{camera\_2} in our experiments).
Estimated camera views after alignment are shown in \cref{fig:app_colmap}.

If COLMAP failed to identify the pose of either of these images, we label scene reconstruction as unsuccessful.
The remainder of evaluation follows the protocol applied for RUST EPE.

\vspace{-2mm}
\paragraph{GNeRF}
\label{app:sec:gnerf}
We use the official GNeRF implementation from the authors~\cite{Meng21iccv_GNeRF}.
We train GNeRF on 3 different scenes of 200 images each, generated from the
MSN dataset.
We select 150 or 12 views views for training and 50 views for evaluation.
GNeRF is trained on each scene independently.

\begin{figure}[t]
\centering
\includegraphics[trim={2cm 15.5cm 2cm 2cm},clip,width=\columnwidth]{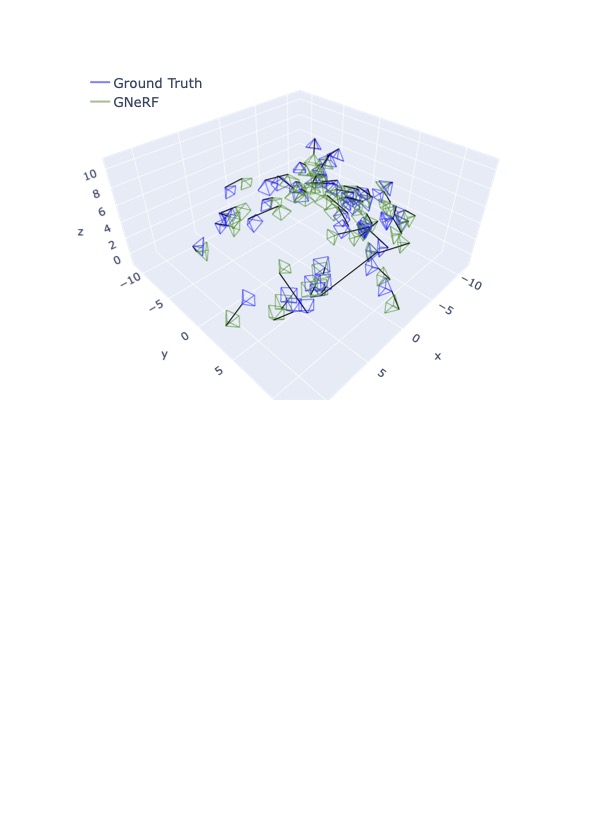}
\caption{
\textbf{GNeRF} --
Camera positions recovered by GNeRF (oriented towards zero-center solely for visualization).
GNeRF predictions (green) approximately recover many ground truth (blue) camera positions, but frequently make very large errors, leading to significant deviations in camera positions.
}
\label{fig:app:gnerf}
\end{figure}

Similar to our evaluation protocol for COLMAP, we transform the predicted camera locations into the coordinate system of the ground-truth camera origins via a rigid transformation (rotation + translation) that minimizes the squared error between camera pairs.
Estimated camera poses after alignment are shown in \cref{fig:app:gnerf}.

To measure MSE and $R^2$, we consider the relative position difference between all camera pairs of the 50 evaluation views, resulting in a total of 2450 pairs for metric computation.
We show GNeRF results on further scenes in \cref{fig:app:qualitative_gnerf}.

\def \y {\textcolor{green}\checkmark}
\def \n {\textcolor{red}\crossmark}

\begin{table*}[t]
\centering
\begin{tabular}{lcllll}
\toprule
Method \quad\:\:\,\:\! & Pose & PSNR $\uparrow$ & SSIM $\downarrow$ & LPIPS $\downarrow$ & LPIPS $\downarrow$ \\
       &      &                 &                   & (AlexNet)          & (VGG) \\
\midrule
SRT~\cite{srt} & \ipgood,\,\tpgood & 23.31 & 0.690 & 0.290 &   0.364 \\
\oursrt & \ipgood,\,\tpgood & 24.40 & 0.740 & 0.237 & 0.314 \\
\oursrt & \ipbad,\,\tpgood & 23.81 & 0.718 & 0.265 & 0.332 \\
\ourupsrt & \ipnone,\,\tpgood & 23.03 & 0.683 & 0.300 & 0.362 \\
\oursrt & \ipbad,\,\tpbad & 18.65 & 0.469 & 0.741 & 0.612 \\
\ourupsrt & \ipnone,\,\tpbad & 18.64 & 0.469 & 0.741 & 0.612 \\
\rust\ & \ipnone,\,\tpnone & 23.49 & 0.703 & 0.287 & 0.351 \\
\bottomrule
\end{tabular}

\begin{tabular}{lcllll}
\toprule
Ablation & \qquad\, & PSNR $\uparrow$ & SSIM $\downarrow$ & LPIPS $\downarrow$ & LPIPS $\downarrow$ \\
         &  &                &                   & (AlexNet)          & (VGG)  \\
\midrule
Right-half PE & & 23.88 & 0.711 & 0.277 & 0.346 \\
Stop grad.& & 23.16 & 0.690 & 0.306 & 0.362 \\
No SLSR & & 20.83 & 0.581 & 0.476 & 0.475 \\
No self-attn & & 22.97 & 0.678 & 0.320 & 0.377 \\
3-dim. \latentpose\ & & 20.40 & 0.563 & 0.506 & 0.492 \\
64-dim. \latentpose\ & & 23.40 & 0.699 & 0.289 & 0.355 \\
768-dim. \latentpose\ & &  23.11 & 0.684 & 0.307 & 0.371 \\
\bottomrule
\end{tabular}
\caption{
    \textbf{Full quantitative results on MSN} -- This table is identical to \cref{tab:psnr}, but additionally reports SSIM~\cite{ssim} and LPIPS~\cite{lpips}.
}
\label{tab:app:moremetrics}
\end{table*}

\subsection{Quantitative and qualitative results [\cref{sec:exp:nvs,sec:exp:sv}]}
\label{app:sec:qualitative}
\label{app:sec:quantitative}

\Cref{app:tab:sv_psnr} summarizes the quantitative results on the Street View dataset from \cref{sec:exp:sv}.
\Cref{fig:app:qualitative_msn,fig:app:qualitative_sv} show more qualitative results on the MSN and SV datasets, respectively.

\begin{figure}[b]
\centering
\includegraphics[width=1.0\columnwidth]{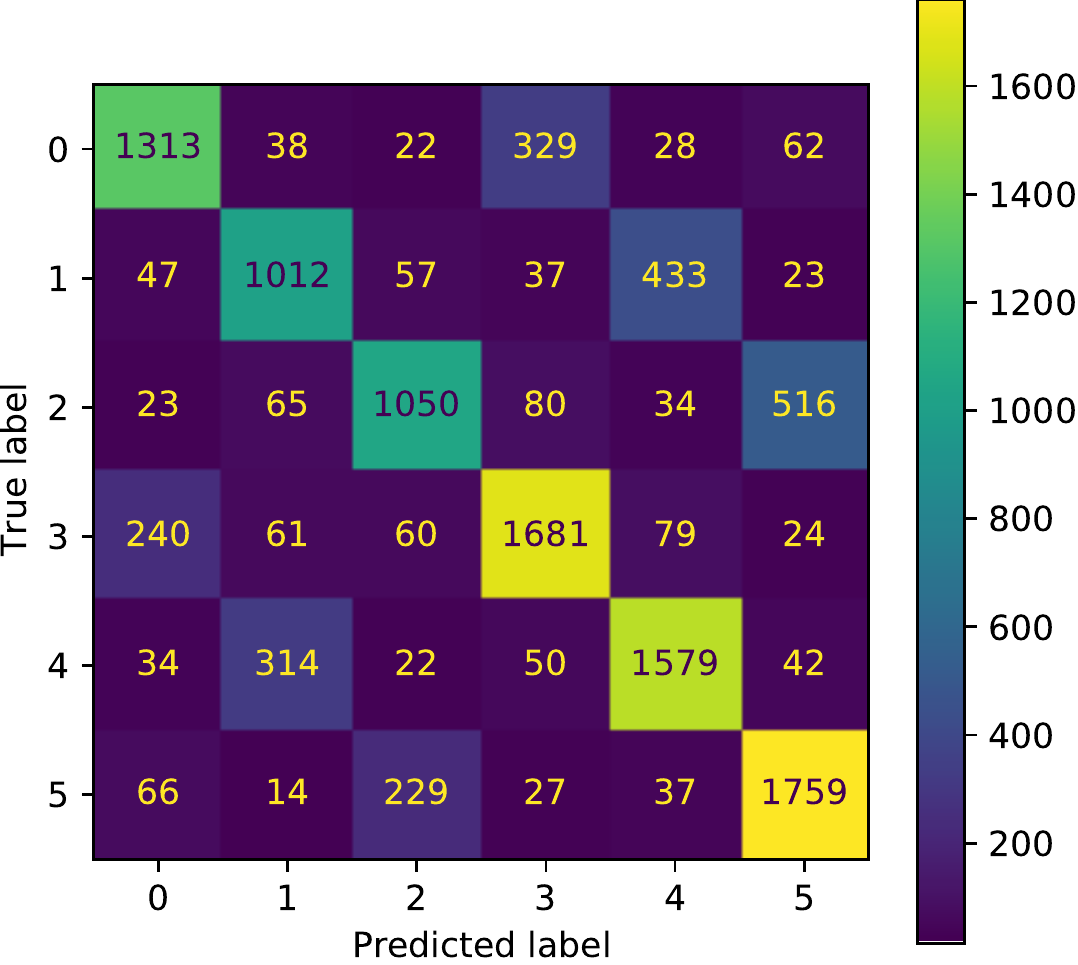}
\caption{
    \textbf{Street View camera ids in latent space} --
    Confusion matrix between the true camera ids (6 cameras on the rig) and the predicted labels of a linear readout from the \rust\ latent pose features \latentpose.
    We can see that the identity is largely predicted correctly.
    Remarkably, the latent pose even distinguishes between opposite cameras, which is much more challenging due to symmetries in the dataset:
    For example, the left-back camera will produce very similar photos to the front-right camera when the car is passing the same street in the opposite direction.
    See \cref{sec:exp:sv} for context.
}
\label{app:fig:confusion}
\end{figure}

\begin{figure}[t!]
    \centering
    \includegraphics[width=0.8\columnwidth]{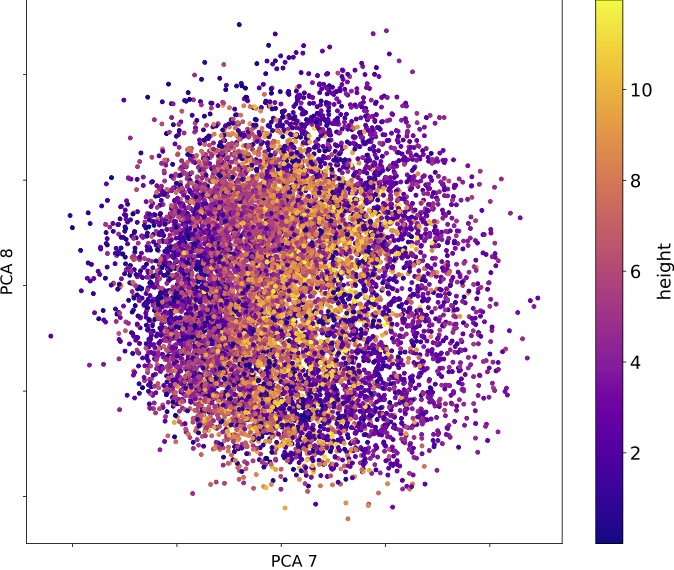}
    \caption{
    \textbf{Latent poses on MSN (cont.)} --
    The seventh and eighth PCA components of the latent poses show some (non-linear) relation with the ground truth camera height.
    }
    \label{fig:app:msn:pca78}
\end{figure}

\begin{table}[b]

\centering
\begin{tabular}{lcc}
\toprule
Model      &    Pose    &       PSNR \\
\midrule
SRT~\cite{srt} & \ipgood,\,\tpgood & 22.72 \\
\oursrt        & \ipgood,\,\tpgood & 23.63 \\
\ourupsrt      & \ipnone,\,\tpgood & 21.25 \\
\rust          & \ipnone,\,\tpnone & 22.50 \\
\bottomrule
\end{tabular}
\caption{
    \textbf{Quantitative results on SV} --
    See \cref{sec:exp:sv} for context.
}
\label{app:tab:sv_psnr}
\end{table}

\newcommand\msntrgrotgraphics[1]{\includegraphics[width=0.094\linewidth]{imgs/appendix/msntr2/cam_360_#1}}  %

\begin{figure*}[t]
    \centering
    \begin{minipage}{0.28\linewidth}
        \includegraphics[width=\linewidth]{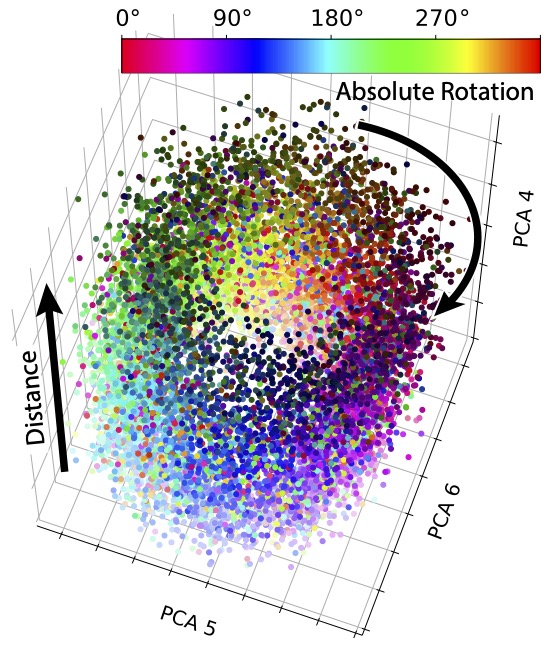}
    \end{minipage}
    \:\:\:\!
    \begin{minipage}{.7\textwidth}
        \centering 
        \setlength{\tabcolsep}{0.2mm}
        \def\arraystretch{2}
        \begin{tabular}{cccccccccc}
            \multicolumn{10}{c}{Global $360^\circ$ rotation}
            \\[-2mm]
            \cmidrule(lr){1-10}
            \\[-8mm]
            \msntrgrotgraphics{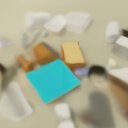} &
            \msntrgrotgraphics{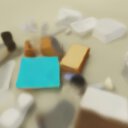} &
            \msntrgrotgraphics{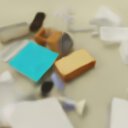} &
            \msntrgrotgraphics{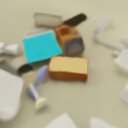} &
            \msntrgrotgraphics{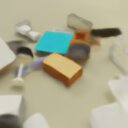} &
            \msntrgrotgraphics{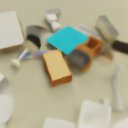} &
            \msntrgrotgraphics{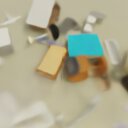} &
            \msntrgrotgraphics{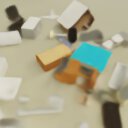} &
            \msntrgrotgraphics{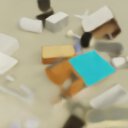} &
            \msntrgrotgraphics{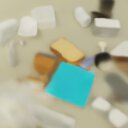}
            \\ [-2mm]
            \msntrgrotgraphics{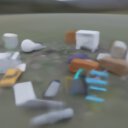} &
            \msntrgrotgraphics{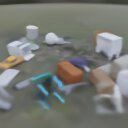} &
            \msntrgrotgraphics{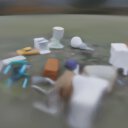} &
            \msntrgrotgraphics{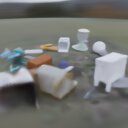} &
            \msntrgrotgraphics{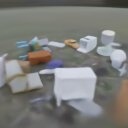} &
            \msntrgrotgraphics{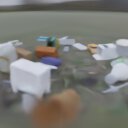} &
            \msntrgrotgraphics{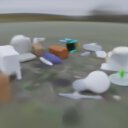} &
            \msntrgrotgraphics{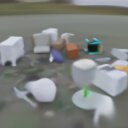} &
            \msntrgrotgraphics{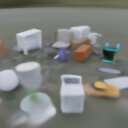} &
            \msntrgrotgraphics{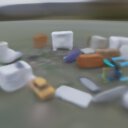}
            \\ [-2mm]
            \msntrgrotgraphics{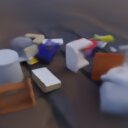} &
            \msntrgrotgraphics{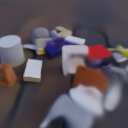} &
            \msntrgrotgraphics{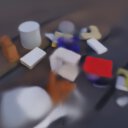} &
            \msntrgrotgraphics{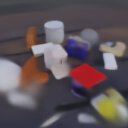} &
            \msntrgrotgraphics{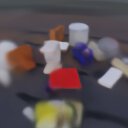} &
            \msntrgrotgraphics{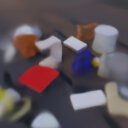} &
            \msntrgrotgraphics{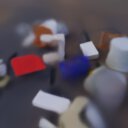} &
            \msntrgrotgraphics{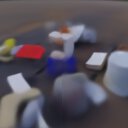} &
            \msntrgrotgraphics{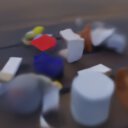} &
            \msntrgrotgraphics{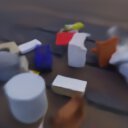}
            \\ [-2mm]
            \msntrgrotgraphics{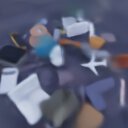} &
            \msntrgrotgraphics{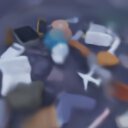} &
            \msntrgrotgraphics{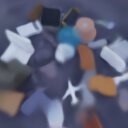} &
            \msntrgrotgraphics{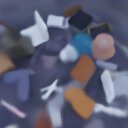} &
            \msntrgrotgraphics{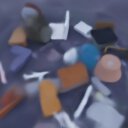} &
            \msntrgrotgraphics{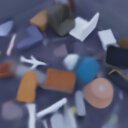} &
            \msntrgrotgraphics{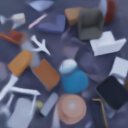} &
            \msntrgrotgraphics{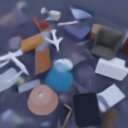} &
            \msntrgrotgraphics{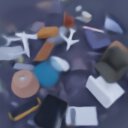} &
            \msntrgrotgraphics{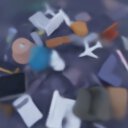}
        \end{tabular}
    \end{minipage}
    \caption{
\textbf{Latent pose investigations on MSN (cont.)} -- In the main manuscript, we investigate the space of latent poses using Principal Component Analysis (PCA).
We found meaningful structure in that the first PCA component encodes height, the second and third encode relative rotation, and the fourth encodes distance from the scene center.
\textbf{Left:} Here we investigate PCA components 5 and 6 which capture absolute rotation.
These components, together with component 4, constitute a cylinder.
Points are coloured such that intensity represents the camera's distance from the scene center and hue represents their rotation around the scene's z-axis in absolute world coordinates.
Rotations around this cylinder's axis are by themselves insufficient to rotate the camera.
Instead, moving $\theta^\circ$ along this cylinder and $3\:\!\theta^\circ$ along the cylinder in PCA components 1, 2 and 3 is necessary to rotate the camera $\theta^\circ$ around the scene's z-axis.
As shown in the \textbf{right}, one can do full $360^\circ$ rotations using this method.
Camera paths that vary distance from the scene center are shown in \cref{fig:msn:traversals} of the main manuscript.
Videos are provided in the included html file.
    }
    \label{fig:app:msn:pca56}
\end{figure*}

\newcommand\msntrablgraphics[2]{\includegraphics[width=0.08\linewidth]{imgs/appendix/msntr2_#2/#1}}

\begin{figure*}[t]
    \centering
    \setlength{\tabcolsep}{0.2mm}
    \def\arraystretch{2}
    \begin{tabular}{cc@{\hskip 0.75mm}cc@{\hskip 0.75mm}cc@{\hskip 1mm}cc@{\hskip 0.75mm}cc@{\hskip 0.75mm}cc}
        \multicolumn{6}{c}{\rust-p64} &
        \multicolumn{6}{c}{\rust-p768}
        \\[-2mm]
        \cmidrule(lr){1-6}
        \cmidrule(lr){7-12}
        \\[-11mm]
        \multicolumn{2}{c}{Height} &
        \multicolumn{2}{c}{Rotation} &
        \multicolumn{2}{c}{Distance} &
        \multicolumn{2}{c}{Height} &
        \multicolumn{2}{c}{Rotation} &
        \multicolumn{2}{c}{Distance}
        \\[-2mm]
        \cmidrule(lr){1-2}
        \cmidrule(lr){3-4}
        \cmidrule(lr){5-6}
        \cmidrule(lr){7-8}
        \cmidrule(lr){9-10}
        \cmidrule(lr){11-12}
        \\[-8mm]
        \msntrablgraphics{cam_height_1_0}{p64} &
        \msntrablgraphics{cam_height_1_63}{p64} &
        \msntrablgraphics{cam_rot_1_0}{p64} &
        \msntrablgraphics{cam_rot_1_62}{p64} &
        \msntrablgraphics{cam_distance_1_0}{p64} &
        \msntrablgraphics{cam_distance_1_63}{p64} &
        \msntrablgraphics{cam_height_1_0}{p768} &
        \msntrablgraphics{cam_height_1_63}{p768} &
        \msntrablgraphics{cam_rot_1_0}{p768} &
        \msntrablgraphics{cam_rot_1_62}{p768} &
        \msntrablgraphics{cam_distance_1_0}{p768} &
        \msntrablgraphics{cam_distance_1_63}{p768}
        \\ [-2mm]
        \msntrablgraphics{cam_height_4_0}{p64} &
        \msntrablgraphics{cam_height_4_63}{p64} &
        \msntrablgraphics{cam_rot_4_0}{p64} &
        \msntrablgraphics{cam_rot_4_62}{p64} &
        \msntrablgraphics{cam_distance_4_0}{p64} &
        \msntrablgraphics{cam_distance_4_63}{p64} &
        \msntrablgraphics{cam_height_4_0}{p768} &
        \msntrablgraphics{cam_height_4_63}{p768} &
        \msntrablgraphics{cam_rot_4_0}{p768} &
        \msntrablgraphics{cam_rot_4_62}{p768} &
        \msntrablgraphics{cam_distance_4_0}{p768} &
        \msntrablgraphics{cam_distance_4_63}{p768}
    \end{tabular}

    \caption{
    \textbf{Latent pose traversals for \rust\ with large latent poses} --
Here we analyze the amount of control afforded by ablations, namely variations of \rust\ that use 64 or 768 dimensions for the latent pose \latentpose.
We refer to these as \rust-p64 and \rust-p768, respectively.
We find that camera motion based on PCA decomposition is only possible to a limited extent for these models.
For example, we found no good way to perform full $360^\circ$ rotations.
Traversals for camera height, distance, and local rotations are feasible. However, these ablations that use higher-dimensional latent poses \latentpose\ are prone to auto-encoding the half of the target image that is input to the Pose Estimator (always the left half here).
Although the right halves of these images demonstrate 3D camera movement, the left halves show wrong content with heavy artifacts.
Videos are provided in the included html file.
    }
    \label{fig:app:msn:ablationtraversals}
\end{figure*}

\begin{figure}[t]
\centering
\includegraphics[trim={2cm 162mm 2cm 18mm},clip,width=\columnwidth]{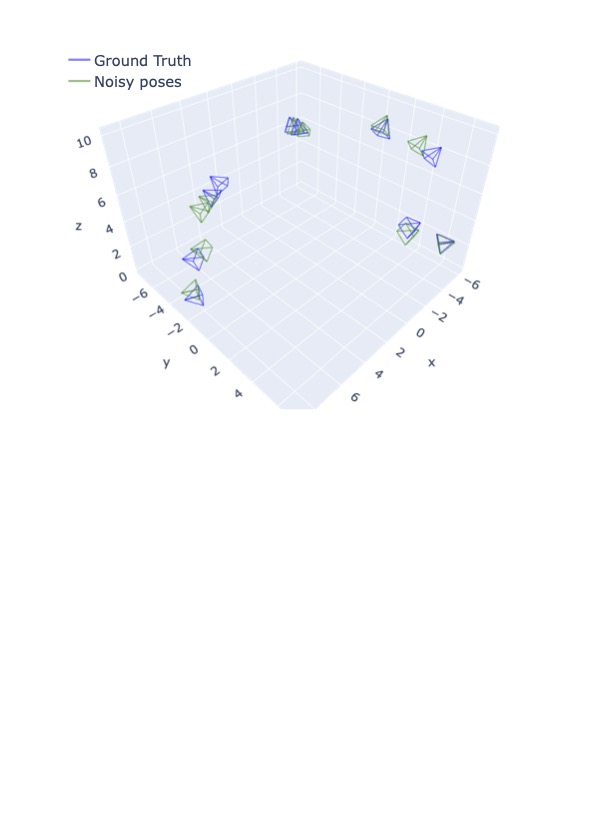}
\caption{
    \textbf{Perturbed camera poses} --
    We visualize the synthetically perturbed camera poses on the MSN dataset with $\sigma$\,$=$\,$0.1$.
    See \cref{sec:exp:nvs} for the performance of the models on noisy cameras.
}
\label{fig:app:camnoise}
\end{figure}

\clearpage

\newcommand\bpsX{0.09}
\newcommand\imggnerfqual[1]{
    \includegraphics[width=\bpsX\linewidth, height=\bpsX\linewidth]{imgs/appendix/gnerf_qual/#1}
}

\newcommand\gerfqualrow[2]{

    \imggnerfqual{ex#1_GT_wBG_#2} &
    \imggnerfqual{ex#1_GT_noBG_#2} &

    \imggnerfqual{ex#1_150im_GT_wBG_#2} &
    \imggnerfqual{ex#1_150im_GT_noBG_#2} &
    \imggnerfqual{ex#1_12im_GT_wBG_#2} &
    \imggnerfqual{ex#1_12im_GT_noBG_#2} &
    
    \imggnerfqual{ex#1_150im__wBG_#2} &
    \imggnerfqual{ex#1_150im__noBG_#2} &
    \imggnerfqual{ex#1_12im__wBG_#2} &
    \imggnerfqual{ex#1_12im__noBG_#2}
    
}

\begin{figure*}[t]
    \centering
    \setlength{\tabcolsep}{-.1mm}
    \def\arraystretch{0.1}
    \begin{tabular}{cc@{\hskip 1mm}cc@{\hskip 1mm}cc@{\hskip 1mm}cc@{\hskip 1mm}cc}
        \multicolumn{2}{c}{Target} &
        \multicolumn{4}{c}{Perfect poses} &
        \multicolumn{4}{c}{No poses}
        \\
        \cmidrule(lr){1-2}
        \cmidrule(lr){3-6}
        \cmidrule(lr){7-10}
        \\
        \multicolumn{2}{c}{} &
        \multicolumn{2}{c}{150 views} &
        \multicolumn{2}{c}{12 views} &
        \multicolumn{2}{c}{150 views} &
        \multicolumn{2}{c}{12 views}
        \\
        \cmidrule(lr){3-4}
        \cmidrule(lr){5-6}
        \cmidrule(lr){7-8}
        \cmidrule(lr){9-10}
        \\
        Normal &
        FG &
        Normal &
        FG &
        Normal &
        FG &
        Normal &
        FG &
        Normal &
        FG
        \\[2mm]
        \gerfqualrow{07}{0}\\[0.5mm]
        \gerfqualrow{07}{1}\\[0.5mm]
        \gerfqualrow{07}{2}\\[2mm]
        PSRN &  & 21.49 & 23.25 & 13.38 & 17.91 & 16.14 & 16.04 & 11.45 & 12.74 \\[1mm]
        $R^2$ &  & & & & & 85.34 & 89.33 & 49.14 & 53.39 \\[1mm]
        MSE &  & & & & & 8.42 & 6.68 & 28.55 & 26.64 \\[8mm]

        \gerfqualrow{10}{0}\\[0.5mm]
        \gerfqualrow{10}{1}\\[0.5mm]
        \gerfqualrow{10}{2}\\[2mm]
        PSNR &  & 24.69 & 24.71 & 17.15 & 20.57 & 18.03 & 19.39 & 13.58 & 12.63 \\[1mm]
         $R^2$ &  &  &  &  &  & 93.66 & 86.88 & 68.91 & 45.52 \\[1mm]
        MSE  & &  &  &  &  & 3.23 & 7.89 & 20.20 & 37.74 \\[8mm]
        \gerfqualrow{11}{0}\\[0.5mm]
        \gerfqualrow{11}{1}\\[0.5mm]
        \gerfqualrow{11}{2}\\[2mm]
        PSNR & & 26.38 & 23.68 & 18.06 & 20.17 & 21.71 & 19.91 & 17.81 & 16.36 \\[1mm]
        $R^2$ & &  &  &  &  & 96.72 & 87.19 & 80.70 & 90.58 \\[1mm]
        MSE & &  &  &  &  & 1.25 & 7.27 & 11.76 & 4.72
    \end{tabular}
    \caption{
        \textbf{More results on GNeRF} --
        Qualitative and quantitative GNeRF~\cite{Meng21iccv_GNeRF} results on further scenes.
        We notice a large variance in terms of model performance.
        See also \cref{sec:exp:epe}.
    }
    \label{fig:app:qualitative_gnerf}
\end{figure*}

\clearpage

\newcommand\bps{0.062}
\newcommand\bp[1]{
    \includegraphics[width=\bps\linewidth, height=\bps\linewidth]{imgs/appendix/qualitative_msn/#1}
}

\begin{sidewaysfigure}[t]
    \vspace{-90mm}
    \setlength{\tabcolsep}{-.1mm}
    \def\arraystretch{0.1}
    \begin{tabular}{c@{\hskip 1mm}c@{\hskip 1mm}ccc@{\hskip 1mm}cc@{\hskip 1mm}c@{\hskip 1mm}ccccccc}
        \multicolumn{1}{c}{Target} &
        \multicolumn{1}{c}{SRT~\cite{srt}} &
        \multicolumn{3}{c}{\oursrt} &
        \multicolumn{2}{c}{\ourupsrt} &
        \multicolumn{1}{c}{\rust} &
        \multicolumn{7}{c}{\rust\ Ablations (\ipnone,\,\tpnone)}
        \\
        \cmidrule(lr){1-1}
        \cmidrule(lr){2-2}
        \cmidrule(lr){3-5}
        \cmidrule(lr){6-7}
        \cmidrule(lr){8-8}
        \cmidrule(lr){9-15}
        \\
        \multicolumn{1}{c}{} &
        \multicolumn{1}{c}{\ipgood,\,\tpgood} &
        \multicolumn{1}{c}{\ipgood,\,\tpgood} &
        \multicolumn{1}{c}{\ipbad,\,\tpgood} &
        \multicolumn{1}{c}{\ipbad,\,\tpbad} &
        \multicolumn{1}{c}{\ipnone,\,\tpgood} &
        \multicolumn{1}{c}{\ipnone,\,\tpbad} &
        \multicolumn{1}{c}{\ipnone,\,\tpnone} &
        \multicolumn{1}{c}{\small Right PE} &
        \multicolumn{1}{c}{\small Stop-grad} &
        \multicolumn{1}{c}{\small No SLSR} &
        \multicolumn{1}{c}{\small No SA} &
        \multicolumn{1}{c}{\small 3-dim \latentpose} &
        \multicolumn{1}{c}{\small 64-dim \latentpose} &
        \multicolumn{1}{c}{\small 768-dim \latentpose}
        \\[2mm]
        \bp{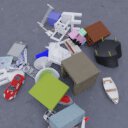} & \bp{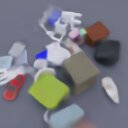} & \bp{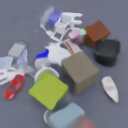} & \bp{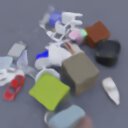} & \bp{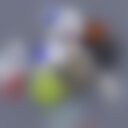} & \bp{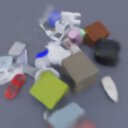} & \bp{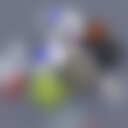} & \bp{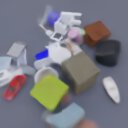} & \bp{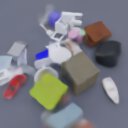} & \bp{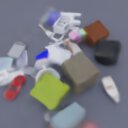} & \bp{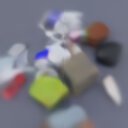} & \bp{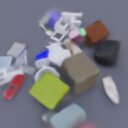} & \bp{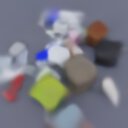} & \bp{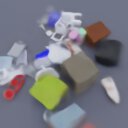} & \bp{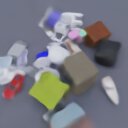}\\[0.5mm]
        \bp{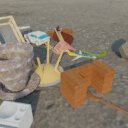} & \bp{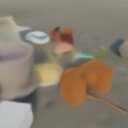} & \bp{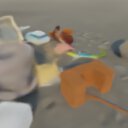} & \bp{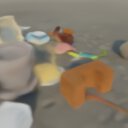} & \bp{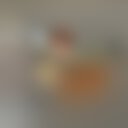} & \bp{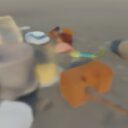} & \bp{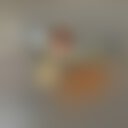} & \bp{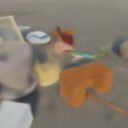} & \bp{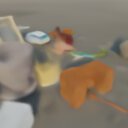} & \bp{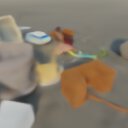} & \bp{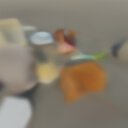} & \bp{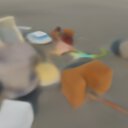} & \bp{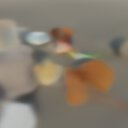} & \bp{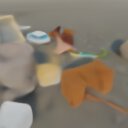} & \bp{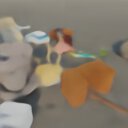}\\[0.5mm]
        \bp{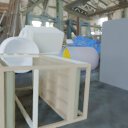} & \bp{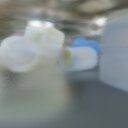} & \bp{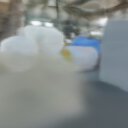} & \bp{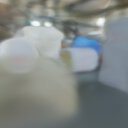} & \bp{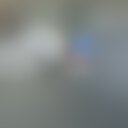} & \bp{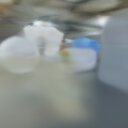} & \bp{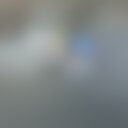} & \bp{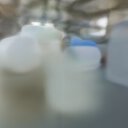} & \bp{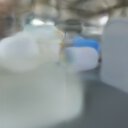} & \bp{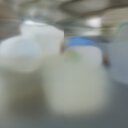} & \bp{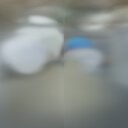} & \bp{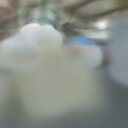} & \bp{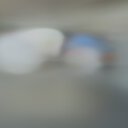} & \bp{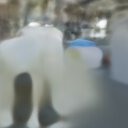} & \bp{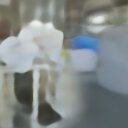}\\[0.5mm]
        \bp{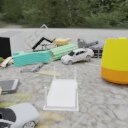} & \bp{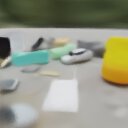} & \bp{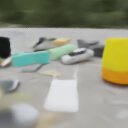} & \bp{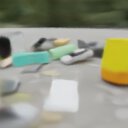} & \bp{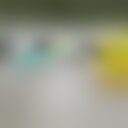} & \bp{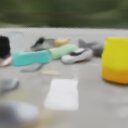} & \bp{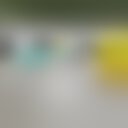} & \bp{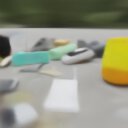} & \bp{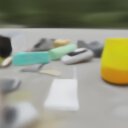} & \bp{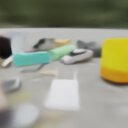} & \bp{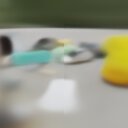} & \bp{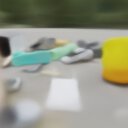} & \bp{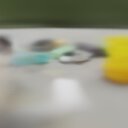} & \bp{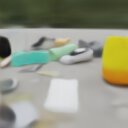} & \bp{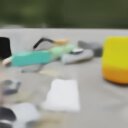}\\[0.5mm]
        \bp{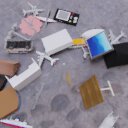} & \bp{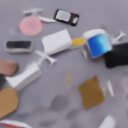} & \bp{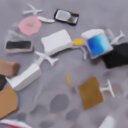} & \bp{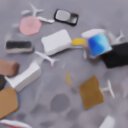} & \bp{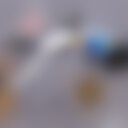} & \bp{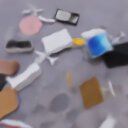} & \bp{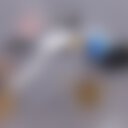} & \bp{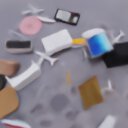} & \bp{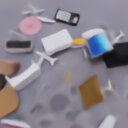} & \bp{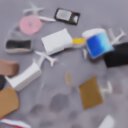} & \bp{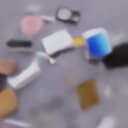} & \bp{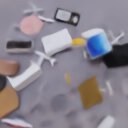} & \bp{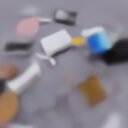} & \bp{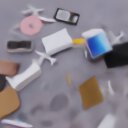} & \bp{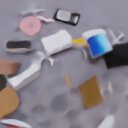}\\[0.5mm]
        \bp{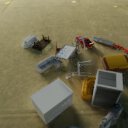} & \bp{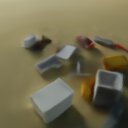} & \bp{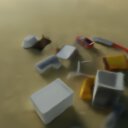} & \bp{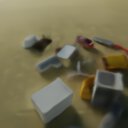} & \bp{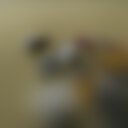} & \bp{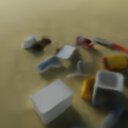} & \bp{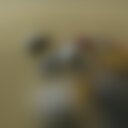} & \bp{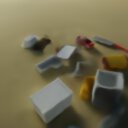} & \bp{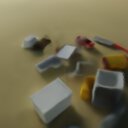} & \bp{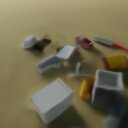} & \bp{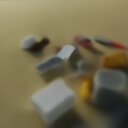} & \bp{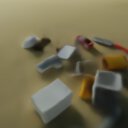} & \bp{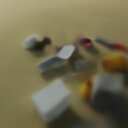} & \bp{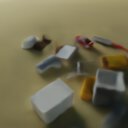} & \bp{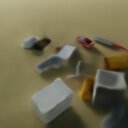}\\[0.5mm]
        \bp{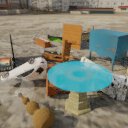} & \bp{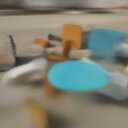} & \bp{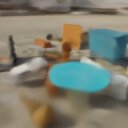} & \bp{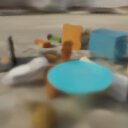} & \bp{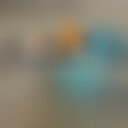} & \bp{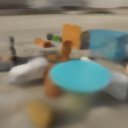} & \bp{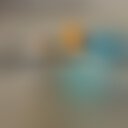} & \bp{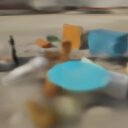} & \bp{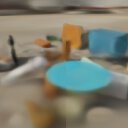} & \bp{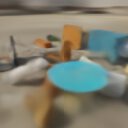} & \bp{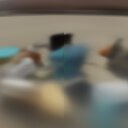} & \bp{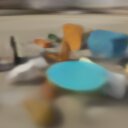} & \bp{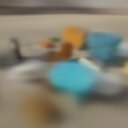} & \bp{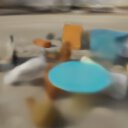} & \bp{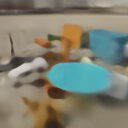}\\[0.5mm]
        \bp{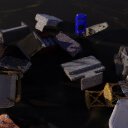} & \bp{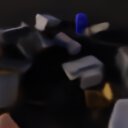} & \bp{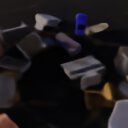} & \bp{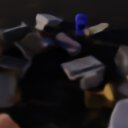} & \bp{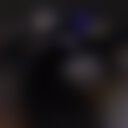} & \bp{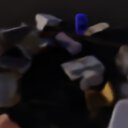} & \bp{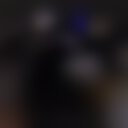} & \bp{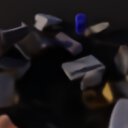} & \bp{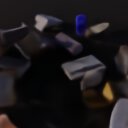} & \bp{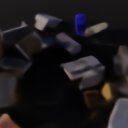} & \bp{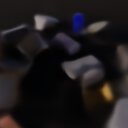} & \bp{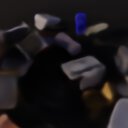} & \bp{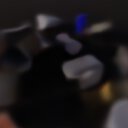} & \bp{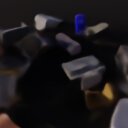} & \bp{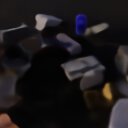}\\[0.5mm]
        \bp{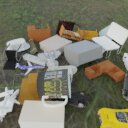} & \bp{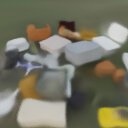} & \bp{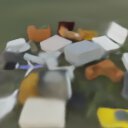} & \bp{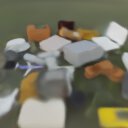} & \bp{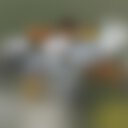} & \bp{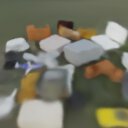} & \bp{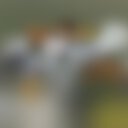} & \bp{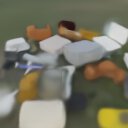} & \bp{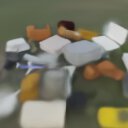} & \bp{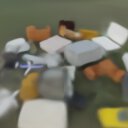} & \bp{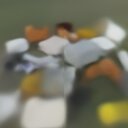} & \bp{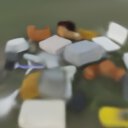} & \bp{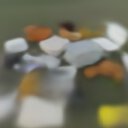} & \bp{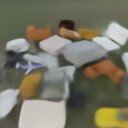} & \bp{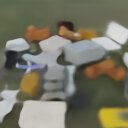}\\[0.5mm]
        \bp{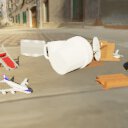} & \bp{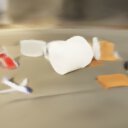} & \bp{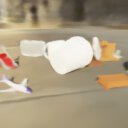} & \bp{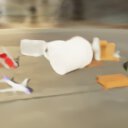} & \bp{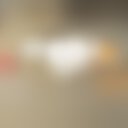} & \bp{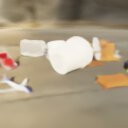} & \bp{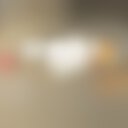} & \bp{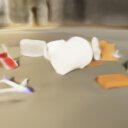} & \bp{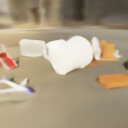} & \bp{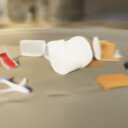} & \bp{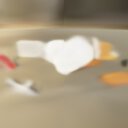} & \bp{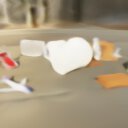} & \bp{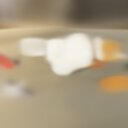} & \bp{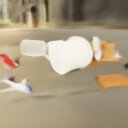} & \bp{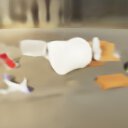}\\[0.5mm]
    \end{tabular}
    \caption{
        \textbf{Full qualitative results on MSN} --
        Left to right:
        Target image,
        SRT~\cite{srt} as originally proposed,
        \oursrt\ with our improved architecture (different camera pose accuracy settings),
        \ourupsrt\ with our improved architecture (different camera pose accuracy settings),
        \rust.
        Further, we show \rust\ ablations (see \cref{sec:exp:ablations}):
        right-half Pose Embedding,
        stop gradient,
        no SLSR,
        no self-attn,
        3-dim. \latentpose,
        64-dim. \latentpose,
        768-dim. \latentpose.
    }
    \label{fig:app:qualitative_msn}
\end{sidewaysfigure}

\clearpage

\newcommand\svps{0.12}
\newcommand\sv[1]{
    \includegraphics[trim={0 0 0 16},clip,height=\svps\linewidth]{imgs/appendix/qualitative_sv/#1}
}

\begin{figure*}[t]
    \centering
    \setlength{\tabcolsep}{-.1mm}
    \def\arraystretch{0.1}
    \begin{tabular}{cc@{\hskip 1mm}cccc}
        Inputs &
        Target &
        SRT~\cite{srt} &
        \oursrt &
        \ourupsrt &
        \rust
        \\
        \sv{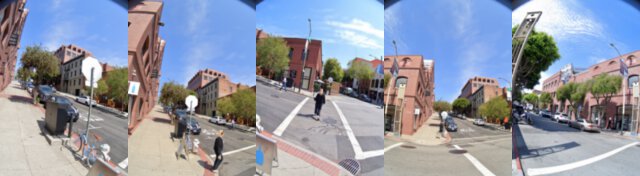} & \sv{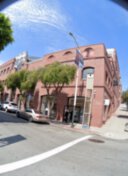} & \sv{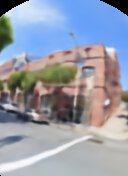} & \sv{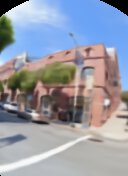} & \sv{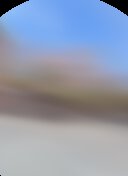} & \sv{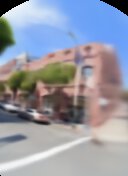}\\[0.5mm]
        \sv{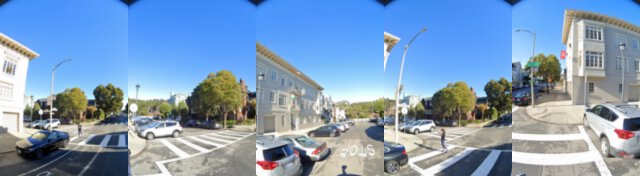} & \sv{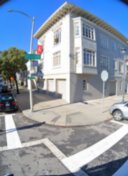} & \sv{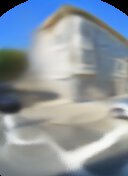} & \sv{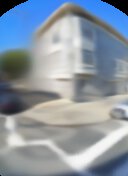} & \sv{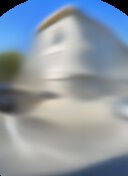} & \sv{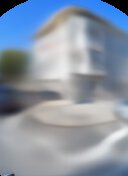}\\[0.5mm]
        \sv{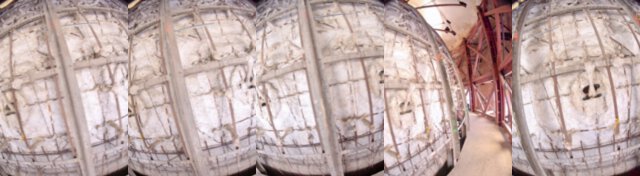} & \sv{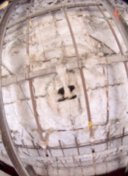} & \sv{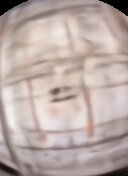} & \sv{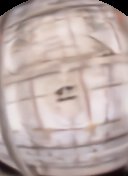} & \sv{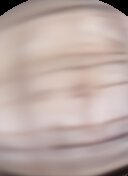} & \sv{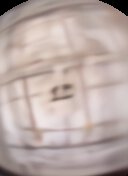}\\[0.5mm]
        \sv{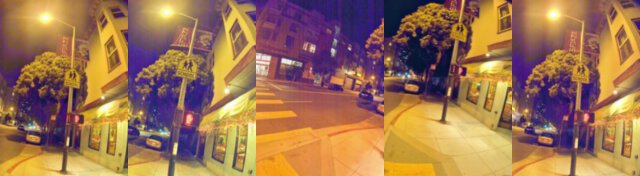} & \sv{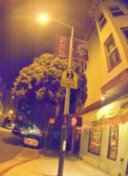} & \sv{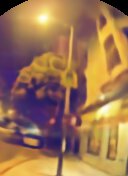} & \sv{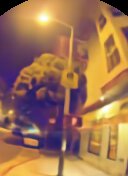} & \sv{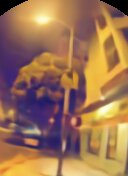} & \sv{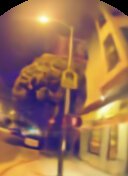}\\[0.5mm]
        \sv{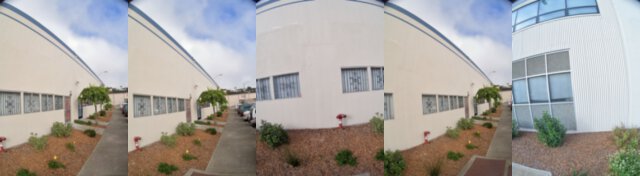} & \sv{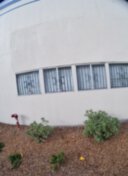} & \sv{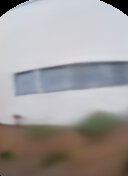} & \sv{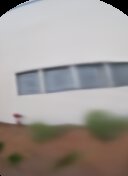} & \sv{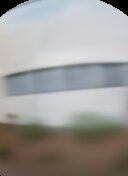} & \sv{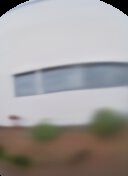}\\[0.5mm]
        \sv{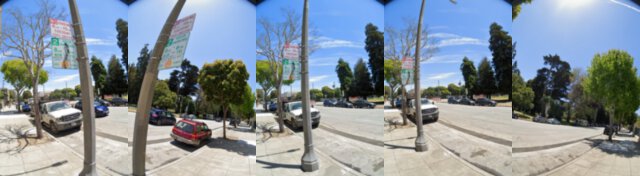} & \sv{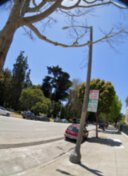} & \sv{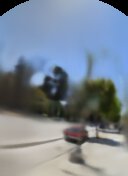} & \sv{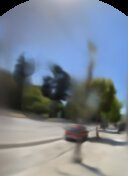} & \sv{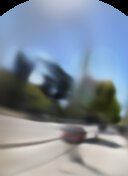} & \sv{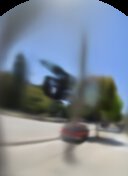}\\[0.5mm]
        \sv{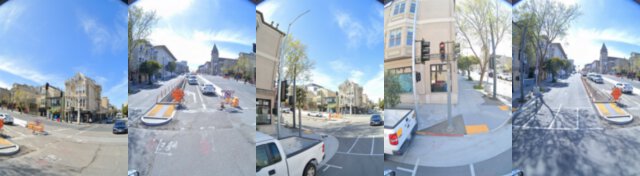} & \sv{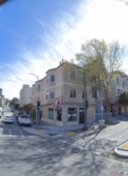} & \sv{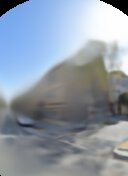} & \sv{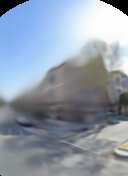} & \sv{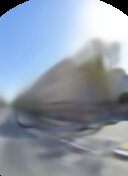} & \sv{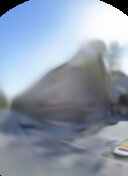}\\[0.5mm]
        \sv{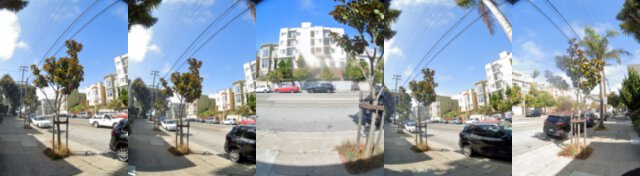} & \sv{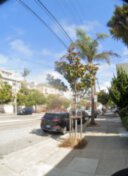} & \sv{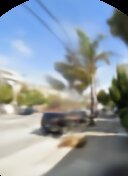} & \sv{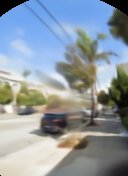} & \sv{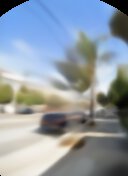} & \sv{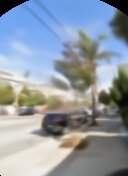}\\[0.5mm]
        \sv{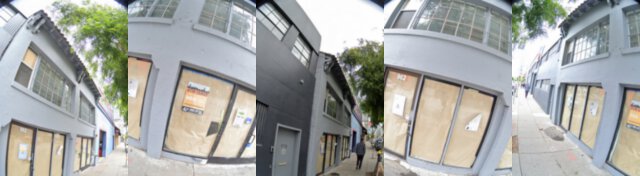} & \sv{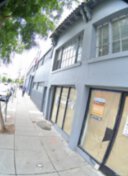} & \sv{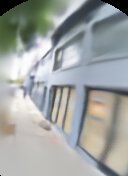} & \sv{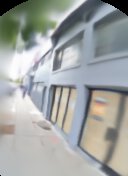} & \sv{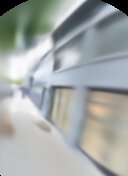} & \sv{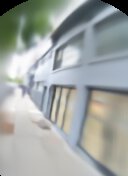}\\[0.5mm]
        \sv{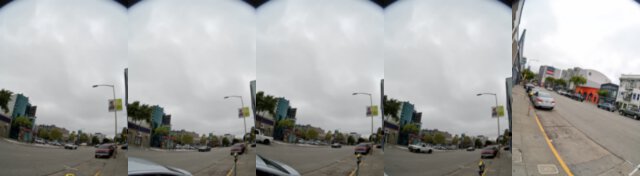} & \sv{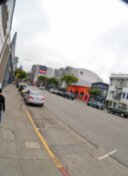} & \sv{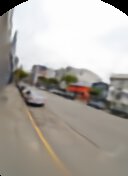} & \sv{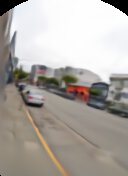} & \sv{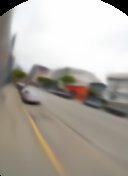} & \sv{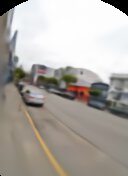}\\[0.5mm]
    \end{tabular}
    \caption{
        \textbf{Full qualitative results on SV} --
        Left to right:
        Target image,
        SRT~\cite{srt} as originally proposed,
        \oursrt\ with our improved architecture,
        \ourupsrt\ with our improved architecture,
        \rust.
        Street View imagery and permission for publication have been obtained from the authors~\cite{GoogleStreetView}.
    }
    \label{fig:app:qualitative_sv}
\end{figure*}

\end{document}